\documentclass[11pt]{article}
\usepackage[margin=1in]{geometry}
\usepackage{amsmath,amssymb,amsthm,mathtools}
\usepackage[T1]{fontenc}
\usepackage{lmodern}
\usepackage{booktabs,longtable,array,graphicx}
\usepackage[hidelinks]{hyperref}
\hypersetup{
  pdftitle={When Parameter Gradient Flows Close in Representation Space},
  pdfauthor={Shengye Tao},
  pdfsubject={Machine learning theory and representation-space gradient flows},
  pdfkeywords={gradient flow, intrinsic dynamics, matrix factorization, conservation laws}
}
\allowdisplaybreaks
\newcommand{\R}{\mathbb R}
\DeclareMathOperator{\rank}{rank}
\DeclareMathOperator{\diag}{diag}
\DeclareMathOperator{\tr}{tr}
\DeclareMathOperator{\grad}{grad}
\DeclareMathOperator{\Span}{span}
\newcommand{\Id}{I}
\newtheorem{theorem}{Theorem}[section]
\newtheorem{lemma}[theorem]{Lemma}
\newtheorem{proposition}[theorem]{Proposition}
\newtheorem{corollary}[theorem]{Corollary}
\theoremstyle{definition}
\newtheorem{definition}[theorem]{Definition}
\theoremstyle{remark}
\newtheorem{remark}[theorem]{Remark}
\title{When Parameter Gradient Flows Close in Representation Space}
\author{Shengye Tao\\[0.35em]
\small Beijing Institute of Mathematical Sciences and Applications (BIMSA)\\
\small Beijing, China\\[0.25em]
\small \href{mailto:ShengyeTao@bimsa.cn}{\texttt{ShengyeTao@bimsa.cn}}\quad
\href{mailto:RanchoTao@gmail.com}{\texttt{RanchoTao@gmail.com}}}
\date{October 20, 2026}
\begin{document}
\maketitle
\begin{abstract}
The represented gradient flow of a parameterized model depends on a mobility
that can retain hidden parameter information. We study when one entire
represented mobility map works for every autonomous loss from one fixed
initialization. For two-factor matrix representations wider than both output
dimensions, with individually full-row-rank factors, we classify all admissible
joint ranks: closure occurs precisely at effective scalar-balanced minimal
joint rank or a two-value signed spectral family at full joint rank. All
intermediate ranks are excluded; positive represented flows belong to known
square-root families. For general representations with constant ambient
representation and gradient-Lie ranks, ordinary closure is equivalent to a
local invariant-level metric in the analytic case and specified smooth
regimes. An endpoint section forces the initially unrestricted mobility map to
admit a locally smooth representative on the represented image. We also separate testing families: linear losses
locally admit closure at every submersion, whereas one fixed translated
strongly convex quadratic family fails it on a delayed-bracket example.
A fixed output readout gives an ordinary squared-loss witness. These
supporting results do not characterize quadratic-only closure in general.
We retain the unresolved smooth output-rank-two case with Lie rank at least
four. A frozen deterministic experiment checks a separate engineered-loss
construction; it is distinguished from the squared-loss proof, universal
theorems and empirical learning performance.
\end{abstract}

\section{Introduction}
For an open parameter domain $\Omega\subset\R^D$ and a smooth representation
$\phi:\Omega\to\R^d$, parameter gradient flow of a represented loss $f$ gives
\begin{equation}\label{eq:flow}
 \dot\theta=-D\phi(\theta)^T\grad f(\phi(\theta)),\qquad
 \dot z=-M(\theta)\grad f(z),\qquad M=D\phi D\phi^T.
\end{equation}
States with the same $z$ may have different $M$. We ask when the entire mobility,
rather than only its action on the gradient of a particular loss, is determined
by $z$ along all loss trajectories from one fixed initialization.
This is the ordinary intrinsic-dynamic contract of \cite{intrinsic},
distinct from its stronger piecewise-loss testing convention.

The two results proved below are an initialization classification for wide
two-factor matrix representations and a converse from ordinary trajectory tests
to invariant-level metric factorization for regular representations in stated
analytic and smooth regimes. All positive matrix mobilities reduce to known
scalar-balanced flows; no new optimizer, optimization rate or task gain is claimed.
The distinction is a necessity theory for eliminating hidden mobility, not a new
formula for the already known positive flow.
The loss-family quantifier is substantive: Proposition~\ref{loss:linear}
shows that all linear tests can pass locally at a submersion, while
Theorem~\ref{loss:quadratic} and Corollary~\ref{loss:squared} give exact
nonclosure with familiar output least-squares losses in a fixed model.
The latter are supporting exclusion controls, not new learning algorithms. Generic continuous initializations
in the nonscalar wide model fail this exact closure, as a direct consequence of
the classification. This does not exclude approximate closure or a simpler
description under one particular loss.

The work uses the conservation-law and Lie-distribution framework of
\cite{conservation,intrinsic}. In particular the gradient Lie algebra and
maximal-stack factor-orbit dimensions in the matrix proof are prior ingredients,
explicitly attributed below. The additional inference under study is how
single-initialization, autonomous-loss tests constrain Gram fibers and a local
invariant-level mobility. Familiar Frobenius, constant-rank, inverse-function,
orbit and ODE results are used as standard local theorems, with their
hypotheses identified at each application.

\paragraph{Results and reading guide.}
Theorem~\ref{mat:classification} gives an initialization classification,
including the distinction between minimal, intermediate, and full joint rank.
Theorem~\ref{bridge:main} supplies the fixed-start converse in its stated regular
regimes. Theorem~\ref{loss:quadratic} separates linear testing from a fixed-Hessian
quadratic family; Corollary~\ref{loss:squared} transfers the witness to squared
error after one fixed readout. The scalar-output case and positive square-root
flows are known. Section~\ref{release:related} compares these statements with
specific predecessor results. Complete proofs are in the appendices.

\section{Problem Setting and Definitions}
Fix $\theta_0\in\Omega$. A solution on an open set means its forward maximal
local interval before exiting that set; no global-time existence is assumed.
Losses are global $\mathcal C^2$ maps $f:\R^d\to\R$ without a common bound on
their derivatives. Neighborhoods can depend on initialization but not on the
tested loss.

\begin{definition}[Ordinary entire-mobility closure]\label{def:ordinary}
Property \textup{(O)} holds at $\theta_0$ if there exist an open neighborhood
$U\ni\theta_0$ and a single map $K:\R^d\to\R^{d\times d}$ such that every
solution of \eqref{eq:flow} from $\theta_0$, for every global $\mathcal C^2$ loss,
satisfies $M(\theta(t))=K(\phi(\theta(t)))$ on its interval in $U$.
The map $K$ is independent of the loss and is initially unrestricted: continuity,
boundedness and differentiability are not assumptions.
\end{definition}

\begin{definition}[Local invariant-level metric]\label{def:metric}
Property \textup{(I)} holds if, on a parameter neighborhood, there are finitely
many common $\mathcal C^1$ conservation laws $h=(h_1,\ldots,h_k)$ and one globally
$\mathcal C^1$ map $\overline K:\R^d\to\R^{d\times d}$ such that
\[
 M(\theta)=\overline K(\phi(\theta))
 \quad\hbox{for every }\theta\in V\cap\{h=h(\theta_0)\}
\]
for some open $V\ni\theta_0$. Common conservation means
$D\phi\,\grad h_j=0$ on the domain of $h$.
\end{definition}

The laws $h$ are local, as in the conservation-law convention of
\cite{intrinsic}; the represented extension $\overline K$ is global.
Equality is required on the actual local level, not merely on its tangent.
Conservation and substitution immediately give \textup{(I)}$\Rightarrow$\textup{(O)}.
The strong intrinsic dynamic property of \cite[Definition~I.2]{intrinsic}
instead tests piecewise $\mathcal C^2$ represented losses $f$, requiring a
continuous trajectory from $\theta_0$ to follow $-\nabla(f\circ\phi)$ wherever
$f$ is differentiable along its represented path, with one common neighborhood
and one common $K$. This is a spatial piecewise-loss convention; arbitrary
time-dependent switching is a separate testing class. Strong closure implies
ordinary closure, so an obstruction to strong closure alone does not prove
failure of ordinary closure. Finite flow concatenations below are auxiliary
geometric constructions; each ordinary witness is realized by an actual
smooth autonomous loss from the fixed initial state.

No positive definiteness on all target space is part of these definitions:
the mobility may be semidefinite when the representation rank is less than $d$.

Let $X_i=\grad\phi_i$ and let $E_\theta$ be the span of the values at $\theta$
of their generated Lie algebra. We use the bracket $[X,Y]=DY\,X-DX\,Y$.
When $\dim E=p$ is constant on an ambient neighborhood, selecting a nonzero
minor of finitely many bracket fields gives a smooth frame for $E$.
The frame brackets lie in $E$, so Frobenius applies. It supplies a local
product box with common first integrals $h$ and a connected initial plaque
$L=\{h=h(\theta_0)\}$ of dimension $p$.
If $p=D$, take $h$ constant. If $\rank D\phi=q$ is constant as well, then
\begin{equation}\label{eq:leafrank}
 \rank D(\phi|_L)=q\le p,
\end{equation}
because $\operatorname{im}D\phi^T\subset T L$ and
$\operatorname{im}(D\phi D\phi^T)=\operatorname{im}D\phi$.
The analytic versions follow by selecting analytic frames and using analytic
Frobenius. These statements concern one local plaque and require neither
globally embedded leaves nor a global quotient.

\begin{lemma}[Compatible loss on an embedded represented arc]\label{lem:jet}
Let a smooth parameter curve satisfy
$\dot\theta=-D\phi(\theta)^Tg(t)$ with smooth $g$, and suppose its represented
curve $\gamma=\phi\circ\theta$ is a regular embedded compact arc admitting a
slightly extended embedded arc. Then this curve is a trajectory of one global
smooth autonomous represented loss from its given initial point.
\end{lemma}
\noindent\textit{Proof.} See Appendix~\ref{release:proof1}.\par

\begin{lemma}[Clock-positive endpoint patch]\label{lem:clock}
Suppose a local embedded manifold $L\ni\theta_0$ is invariant for all fields
$Y_g=-D\phi^Tg$, $g\in\R^d$, and their Lie values span $T_\theta L$ on $L$.
Assume $M(\theta_0)\ne0$. Every sufficiently small ambient neighborhood admits
a nonempty $L$-open set of endpoints from $\theta_0$, each realized by one
global smooth autonomous represented loss with its whole path in that
neighborhood. Consequently \textup{(O)} implies $M=K\circ\phi$ on this set.
Constant Lie rank in the ambient parameter space is not required.
\end{lemma}
\noindent\textit{Proof.} See Appendix~\ref{release:proof2}.\par

\begin{lemma}[Linear losses force local regularity]\label{lem:section}
Suppose $\phi$ is smooth with constant ambient rank $q>0$ near $\theta_0$.
Under \textup{(O)}, its common arbitrary $K$ agrees on the whole local represented
image near $z_0=\phi(\theta_0)$ with a smooth function, admitting a smooth
extension $K_{\rm loc}$ to a target neighborhood. If $\phi$ is analytic the
section and local extension can be analytic. No Lie-rank assumption is needed.
\end{lemma}
\noindent\textit{Proof.} See Appendix~\ref{release:proof3}.\par

\begin{lemma}[Local descent and global represented extension]\label{lem:descent}
Let $L$ be a local plaque on which $\phi$ has constant rank $q$. If
$D(M|_L)$ annihilates $\ker D(\phi|_L)$ throughout a neighborhood of
$\theta_0$ in $L$, then $M=k\circ\phi$ on a smaller plaque neighborhood for a
smooth local represented $k$. If $L$ is the exact local level of common laws,
this gives \textup{(I)} after shrinking the parameter box.
\end{lemma}
\noindent\textit{Proof.} See Appendix~\ref{release:proof4}.\par

\section{Complete classification for wide two-factor representations}
\label{mat:appendix}

We identify matrix spaces with Euclidean spaces using the Frobenius inner
product.  Put
\[
 N=n+m,\qquad J=\diag(\Id_n,-\Id_m),\qquad
 F=\begin{bmatrix}U\\V\end{bmatrix},\qquad
 C=FF^\top=\begin{bmatrix}A&Z\\Z^\top&B\end{bmatrix},
 \qquad S=F^\top JF.
\]
Here \(U\in\R^{n\times r}\), \(V\in\R^{m\times r}\), and
\(\phi(F)=Z=UV^\top\).  Direct differentiation gives
\[
 D\phi(F)[\delta U,\delta V]
       =\delta U V^\top+U\delta V^\top,\qquad
 D\phi(F)^\top G=\begin{bmatrix}GV\\G^\top U\end{bmatrix}.
\]
Consequently the full represented mobility is the linear operator
\begin{equation}
 M_F[G]=AG+GB,\qquad A=UU^\top,\quad B=VV^\top.
 \label{mat:mobility}
\end{equation}
For \(G\in\R^{n\times m}\), write
\[
 H_G=\begin{bmatrix}0&G\\G^\top&0\end{bmatrix},
 \qquad X_G(F)=-H_GF.
\]
These are the parameter gradient fields of the linear represented losses.
An arbitrary represented loss \(f\in C^2(\R^{n\times m},\R)\) gives
\begin{equation}
 \dot F=-H_{\grad f(Z)}F.
 \label{mat:factorflow}
\end{equation}

For clarity, property \(\textup{(O)}\) at \(F_0\) means that one map
\(K:\R^{n\times m}\to
\R^{(nm)\times(nm)}\), independent of \(f\), equals the entire \(M_F\)
along every such autonomous flow from \(F_0\), until first exit from
one common parameter neighborhood.  No regularity of \(K\) is assumed.
Property \(\textup{(I)}\) means that a finite family of common local
\(C^1\) conservation laws \(h\), together with a globally \(C^1\)
represented map \(\overline K\), gives \(M_F=\overline K(\phi(F))\)
on the exact local level \(h(F)=h(F_0)\).  The conservation laws need
only be defined on a parameter neighborhood.  Conservation and the
chain rule show that \(\textup{(I)}\) implies \(\textup{(O)}\).

The gradient Lie algebra and the factor-orbit dimensions used below
are prior ingredients of the conservation-law framework of Marcotte,
Gribonval, and Peyr\'e \cite{conservation}, Propositions H.3 and H.5.  We include their elementary
linear-algebra derivations to make the present argument complete.
Likewise the positive scalar-balanced square-root dynamics are known \cite{intrinsic,domine};
the classification below concerns their initialization conditions.

\begin{theorem}[Wide initialization classification]
\label{mat:classification}
Assume \(n,m\geq1\), \(r>\max(n,m)\), and
\(\rank U_0=n\), \(\rank V_0=m\).
Set \(L_0=\max(n,m)\), \(s=\rank F_0\), and \(C_0=F_0F_0^\top\).
Properties \(\textup{(O)}\) and \(\textup{(I)}\), with the latter
interpreted in the original width-\(r\) parameter space, are equivalent
to the union of the following disjoint cases:
\begin{align}
 &s=L_0,\qquad C_0JC_0=\lambda C_0
                  \quad\hbox{for some }\lambda\in\R; \label{mat:mincase}\\
 &s=N,\qquad \operatorname{spec}(JC_0)
       =\{a\text{ with multiplicity }n,\,
                    -b\text{ with multiplicity }m\},
       \quad a,b>0. \label{mat:fullcase}
\end{align}
In \eqref{mat:mincase}, \(n>m\) forces \(\lambda>0\), \(m>n\)
forces \(\lambda<0\), and \(n=m\) permits any \(\lambda\).
There are no positive cases with \(L_0<s<N\).
The mobility in either positive case is
\begin{equation}
 K_\sigma(Z)[G]
  =\sqrt{\sigma^2\Id_n+ZZ^\top}\,G
       +G\sqrt{\sigma^2\Id_m+Z^\top Z},
 \label{mat:closedmobility}
\end{equation}
where \(\sigma=|\lambda|/2\) in \eqref{mat:mincase} and
\(\sigma=(a+b)/2\) in \eqref{mat:fullcase}.
For \(\sigma=0\), the formula is used on an invertible represented
neighborhood and extended smoothly by a cutoff.
\end{theorem}

\begin{figure}[tbp]
\centering
\includegraphics[width=\textwidth]{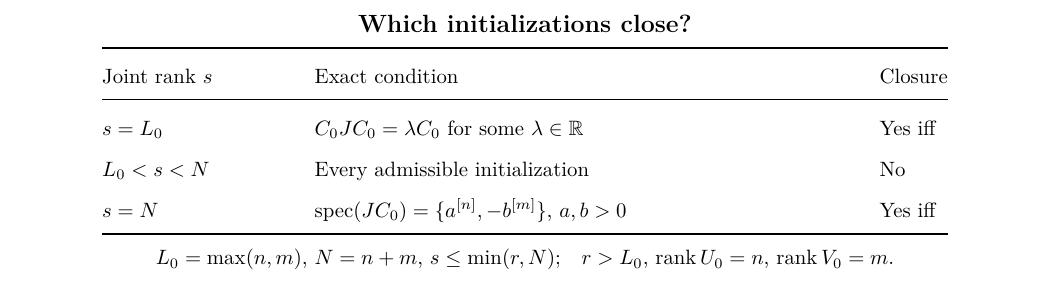}
\caption{Initialization classification in Theorem~\ref{mat:classification}.
Here $L_0=\max(n,m)$, $N=n+m$, $s=\rank F_0$, $C_0=F_0F_0^\top$,
and $J=\diag(I_n,-I_m)$. Only ranks $s\leq r$ are attainable.
The assumptions $r>L_0$, $\rank U_0=n$, and $\rank V_0=m$ apply to every row.
At minimal joint rank, $n>m$ forces $\lambda>0$ and $m>n$ forces
$\lambda<0$; when $n=m$, either sign or zero is allowed.
This is a theorem summary, not a numerical experiment.}
\label{fig:classification}
\end{figure}
\subsection{Factor orbits and the ordinary-closure obstruction}

\begin{lemma}[Boost algebra and factor-orbit dimension]
\label{mat:factororbit}
The fields \(X_G\) generate the left action of
\[
 \mathfrak o(n,m)=\{T\in\R^{N\times N}:T^\top J+JT=0\}.
\]
At a factor matrix of rank \(s\), its local orbit has dimension
\begin{equation}
 d_F=\binom N2-\binom{N-s}2
       =\frac{s(2N-s-1)}2. \label{mat:factordim}
\end{equation}
This formula does not require \(J\) restricted to \(\operatorname{im}F\)
to be nondegenerate.  Every common \(C^1\) conservation law is constant
on the connected local orbit.
\end{lemma}
\noindent\textit{Proof.} See Appendix~\ref{release:proof5}.\par

\begin{lemma}[A necessary Gram-orbit dimension bound]
\label{mat:gramobstruction}
If \(F_0\ne0\) satisfies \(\textup{(O)}\), then the congruence orbit of
\(C_0=F_0F_0^\top\) under the \(J\)-orthogonal group has dimension
\(d_C\leq nm\).  The same bound follows from \(\textup{(I)}\), for
arbitrary finite families of common \(C^1\) conservation laws.
\end{lemma}
\noindent\textit{Proof.} See Appendix~\ref{release:proof6}.\par

\subsection{The full-column and degenerate-inertia calculations}

\begin{lemma}[Full-column Gram rank]
\label{mat:column}
Suppose \(F\in\R^{N\times s}\) has full column rank, and let
\(c_1,\ldots,c_\ell\) be the multiplicities of all distinct eigenvalues
of the symmetric \(S=F^\top JF\), including zero when present.
Then
\begin{equation}
 d_C=\frac{s(2N-s-1)}2-\sum_{j=1}^{\ell}\binom{c_j}2.
 \label{mat:gramcolumn}
\end{equation}
\end{lemma}
\noindent\textit{Proof.} See Appendix~\ref{release:proof7}.\par

\begin{lemma}[Strict excess at intermediate width]
\label{mat:intermediate}
For \(\max(n,m)<s<N\) and full-column \(F\in\R^{N\times s}\),
one has \(d_C>nm\), whether or not \(S\) is singular.
\end{lemma}
\noindent\textit{Proof.} See Appendix~\ref{release:proof8}.\par

\begin{lemma}[The minimal full-column width]
\label{mat:minimal}
Suppose \(s=L_0=\max(n,m)\), \(F\) is full column rank, and both
blocks have their full row ranks.
Then \(d_C\leq nm\) if and only if \(S=\lambda\Id_s\).
This is equivalent to \(CJC=\lambda C\), with the sign conditions in
Theorem~\ref{mat:classification}.
\end{lemma}
\noindent\textit{Proof.} See Appendix~\ref{release:proof9}.\par

\subsection{Full joint rank and the signed spectral family}

\begin{lemma}[Full-row Gram orbit and its positive family]
\label{mat:row}
Suppose \(F\in\R^{N\times r}\) has full row rank, so \(C=FF^\top>0\).
Let \(c_i\) and \(d_j\) be the multiplicities of the distinct positive
and negative eigenvalue values, respectively, of \(JC\).
Then
\begin{equation}
 d_C-nm
   =\frac{n^2-\sum_i c_i^2+m^2-\sum_j d_j^2}{2}.
 \label{mat:gramrow}
\end{equation}
Thus \(d_C\leq nm\) precisely when \(JC\) has one positive value \(a\)
and one negative value \(-b\), with multiplicities \(n,m\).
At such an initialization the original ambient canonical level
\(F^\top JF=F_0^\top JF_0\) has mobility
\eqref{mat:closedmobility}, with \(\sigma=(a+b)/2>0\).
\end{lemma}
\noindent\textit{Proof.} See Appendix~\ref{release:proof10}.\par

\subsection{Effective hidden width and the original ambient level}

\begin{lemma}[Exact width reduction for fixed-start trajectories]
\label{mat:reduction}
Let \(s=\rank F_0\), and choose an orthogonal hidden matrix \(Q\) with
\(F_0Q=[\widetilde F_0,0]\), where \(\widetilde F_0\) has \(s\)
independent columns.
For every represented \(C^2\) loss, the original flow from \(F_0\)
and the width-\(s\) flow from \(\widetilde F_0\) have identical
\(Z,A,B,M\) on corresponding existence intervals.
Property \(\textup{(O)}\) holds at one of these initializations if and
only if it holds at the other.  If both original blocks have full row
rank, so do the reduced blocks.
\end{lemma}
\noindent\textit{Proof.} See Appendix~\ref{release:proof11}.\par

\begin{lemma}[Minimal-rank sufficiency in the original width]
\label{mat:ambientboundary}
Under the assumptions of Theorem~\ref{mat:classification}, suppose
\(s=L_0\) and \(C_0JC_0=\lambda C_0\).
Then \(\textup{(I)}\) holds in the original width-\(r\) space,
using the canonical common invariant \(h(F)=F^\top JF\).
\end{lemma}
\noindent\textit{Proof.} See Appendix~\ref{release:proof12}.\par

\noindent\textit{Proof of Theorem~\ref{mat:classification}.} See Appendix~\ref{release:proof13}.\par

\subsection{The matching known positive dynamics}

\begin{proposition}[Exact narrow scalar-balanced realization]
\label{mat:knownpositive}
For \(\sigma>0\) and every \(Z\in\R^{n\times m}\),
the operator \eqref{mat:closedmobility} is realized by width
\(\min(n,m)\) factors with scalar hidden imbalance.
This includes rank-deficient \(Z\).
For \(\sigma=0\) and square invertible \(Z\), a balanced realization
exists locally as well.
Thus the positive cases in Theorem~\ref{mat:classification} do not
provide new represented continuous-time trajectories.
\end{proposition}
\noindent\textit{Proof.} See Appendix~\ref{release:proof14}.\par

\begin{corollary}[Generic continuously sampled wide factors]
\label{mat:generic}
Let \(r>\max(n,m)\).
For any jointly absolutely continuous initialization law on the
full factor space, properties \(\textup{(O)}\) and \(\textup{(I)}\)
have probability zero if \((n,m)\ne(1,1)\).
For the scalar pair they have probability one.
No independence of the factor entries is required.
\end{corollary}
\noindent\textit{Proof.} See Appendix~\ref{release:proof15}.\par

\begin{remark}
The corollary is a direct classification consequence.
It concerns exact entire-mobility closure, with an unrestricted
universal loss-testing class.  It gives no quantitative approximate
closure, learning-performance, stochastic, or discrete-time claim.
Neither the classification nor its proof covers every individually
rank-deficient block initialization.
\end{remark}

\section{From initialized trajectory tests to a local metric}
\label{bridge:appendix}

We use the ordinary closure and local metric properties in
Definitions~\ref{def:ordinary} and~\ref{def:metric}.  Throughout this
section the parameter metric is the original Euclidean metric.  Write
\[
 M=D\phi D\phi^{\mathsf T},\qquad
 \mathcal H=\operatorname{image}D\phi^{\mathsf T},\qquad
 \mathcal E=\operatorname{Lie}\{\grad\phi_i:1\leq i\leq d\}.
\]
Here the last expression denotes the pointwise distribution of the
generated Lie algebra.  Unless stated otherwise, its rank $p$ and the
rank $q$ of $D\phi$ are constant on an ambient open neighborhood of the
initialization $\theta_0$.  All subsequent shrinkages are inside the
common neighborhood in the ordinary closure hypothesis.  Frobenius,
the constant-rank theorem, smooth dependence of ODE solutions, and
tubular coordinates are used in their classical local forms.

\begin{lemma}[The regular initial plaque]
\label{bridge:plaque}
There is a smooth Frobenius product box with finitely many smooth common
conservation laws $h$, whose initial level is one connected
$p$-dimensional plaque $L$ tangent to $\mathcal E$.  If $\phi$ is real
analytic, this box, $L$, and $h$ can be chosen real analytic.  On $L$,
$D(\phi|_L)$ has constant rank $q$.
\end{lemma}
\noindent\textit{Proof.} See Appendix~\ref{release:proof16}.\par

\begin{theorem}[Regular analytic and specified smooth regimes]
\label{bridge:main}
For a real-analytic representation with the two constant ambient ranks,
ordinary closure is equivalent to a local invariant-level metric.
The same equivalence holds for a smooth representation when $q\geq3$,
when $q\leq1$, or when $q=2$ and $p\leq3$.  The conservation laws and
the global represented extension in the conclusion can be smooth.
\end{theorem}

\subsection{The analytic case and the exact descent}
\label{bridge:analytic-section}

\noindent\textit{Proof of Theorem~\ref{bridge:main} in the analytic case.} See Appendix~\ref{release:proof17}.\par

\subsection{Smooth horizontal paths in represented rank at least three}

\begin{lemma}[An elementary embedded-arc approximation]
\label{bridge:embedded}
Let $q\geq3$ and let $w:[0,T]\to\R^q$, $T>0$, be smooth, with compact
image in an open coordinate set.  There are vectors $a_j\to0$ for
which $w_j(t)=w(t)+ta_j$ are regular embedded smooth arcs, remain in the
same coordinate set, have $w_j(0)=w(0)$, and converge to $w$ in $C^1$.
Their other endpoints need only converge.
\end{lemma}
\noindent\textit{Proof.} See Appendix~\ref{release:proof18}.\par

\begin{lemma}[Closure on a smooth horizontal path]
\label{bridge:horizontal}
Assume ordinary closure, the constant representation rank, and
$q\geq3$.  On a sufficiently small parameter neighborhood, every
smooth path from $\theta_0$ tangent to $\mathcal H$ satisfies
$M=K_{\mathrm{loc}}\circ\phi$, where $K_{\mathrm{loc}}$ is the forced
smooth representative of Lemma~\ref{lem:section}.
\end{lemma}
\noindent\textit{Proof.} See Appendix~\ref{release:proof19}.\par

\begin{lemma}[Local finite-word accessibility]
\label{bridge:accessibility}
On the regular plaque $L$, every point of a sufficiently small
neighborhood of $\theta_0$ is the endpoint of a finite horizontal
flow word lying in a prescribed slightly larger plaque neighborhood.
Such words admit smooth horizontal time reparametrizations with the
same endpoints.
\end{lemma}
\noindent\textit{Proof.} See Appendix~\ref{release:proof20}.\par

\noindent\textit{Proof of Theorem~\ref{bridge:main} for $q\ge3$ and the elementary low-rank cases.} See Appendix~\ref{release:proof21}.\par

\subsection{The rank-two, three-dimensional-leaf case}

\begin{lemma}[A finite horizontal curvature jet]
\label{bridge:curvature-jet}
Suppose $q=2,p=3$.  On $L$, choose coordinates $(u,v,\eta)$ with
$(u,v)=\pi$ represented coordinates and $\theta_0=(0,0,0)$.  Write its
horizontal lifts as
\[
 H_1=\partial_u+A\partial_\eta,\qquad
 H_2=\partial_v+B\partial_\eta,\qquad
 [H_1,H_2]=c\partial_\eta.
\]
Some finite ordered horizontal derivative of $c$ is nonzero at
$\theta_0$.  If $m$ is the least such order, the leading word tensor
is symmetric and the homogeneous polynomial
$P_m(e)=H_e^m c(\theta_0)$, $H_e=e_1H_1+e_2H_2$, is not identically
zero.  For $m=0$ this means $P_0=c(\theta_0)\ne0$.
\end{lemma}
\noindent\textit{Proof.} See Appendix~\ref{release:proof22}.\par

\begin{lemma}[Four rectangle families and a common fiber interval]
\label{bridge:rectangles}
Under the preceding hypotheses there are two independent unit base
directions and their negatives, and one fixed small $\varepsilon>0$,
such that the signed rectangle loops in any of these four directions
have horizontal endpoints covering one common open interval about
$\theta_0$ in the initial one-dimensional fiber.
\end{lemma}
\noindent\textit{Proof.} See Appendix~\ref{release:proof23}.\par

For the next lemma, let $\mathcal D=\{\pm e^{(1)},\pm e^{(2)}\}$ and
$\varepsilon>0$ be the directions and fixed length supplied by
Lemma~\ref{bridge:rectangles}. For $e\in\mathcal D$, let $e^\perp$ be
its counterclockwise rotation through $\pi/2$. The rectangle of signed
width $\delta$ is the loop
\[
0\longrightarrow\varepsilon e
\longrightarrow\varepsilon e+\delta e^\perp
\longrightarrow\delta e^\perp\longrightarrow0,
\]
with $\delta$ sufficiently small.

\begin{lemma}[The required planar approximation]
\label{bridge:planar}
For any sufficiently small nonzero base point $b$, one may select
$e\in\mathcal D$ with $e\cdot b<0$.  For fixed sufficiently small nonzero $\delta$,
the rectangle in Lemma~\ref{bridge:rectangles}, followed by the
straight tail from zero to $b$, is a limit of regular embedded smooth
base arcs with initial point exactly zero and terminal point $b$.
Their velocities converge in $L^1$ and their total variations remain
bounded.  For $b=0$ their terminal points may instead converge to
zero.  These assertions suffice to pass the corresponding horizontal
lift endpoints through the forced smooth $K_{\mathrm{loc}}$.
\end{lemma}
\noindent\textit{Proof.} See Appendix~\ref{release:proof24}.\par

\noindent\textit{Proof of Theorem~\ref{bridge:main} for $q=2,p=3$.} See Appendix~\ref{release:proof25}.\par

\subsection{All-order and Euclidean necessary conditions}

\begin{proposition}[All-order flatness at the initial plaque]
\label{bridge:flatness}
Under smooth constant ambient ranks and ordinary closure, take
$K_{\mathrm{loc}}=0$ when $q=0$ and otherwise use the smooth representative
from Lemma~\ref{lem:section}. Set
$R=M-K_{\mathrm{loc}}\circ\phi$.  Every finite plaque-coordinate
derivative of $R|_L$ at $\theta_0$ vanishes.
\end{proposition}
\noindent\textit{Proof.} See Appendix~\ref{release:proof26}.\par

\begin{proposition}[A Euclidean output obstruction]
\label{bridge:hessian}
Suppose $q=2$.  Let $u=a\cdot\phi+c_0$ and $v=b\cdot\phi$ be
constant output-linear combinations with independent gradients
$X=\grad u$, $Y=\grad v$ on the chosen neighborhood.  Assume
$D^2u\,Y\in\mathcal H$ there.  Ordinary closure holds if and only
if $p=2$.  In particular this applies to an affine $u$, or to an
independent isotropic quadratic
$u=\alpha\|\theta\|^2/2+a_0\cdot\theta+c_0$ that is such an
output combination.
\end{proposition}
\noindent\textit{Proof.} See Appendix~\ref{release:proof27}.\par

\begin{remark}
\label{bridge:scope}
The general smooth case $q=2,p\geq4$ is not settled here.
Proposition~\ref{bridge:flatness} makes a counterexample's actual
mobility discrepancy flat to every order at the initialization, but
does not rule out a nonzero smooth flat discrepancy.  A planar
non-density observation or an arbitrary connection is not by itself
a Euclidean representation satisfying the universal ordinary
premise.  None of these results asserts a new positive optimizer,
discrete dynamics, or an implication from a single observed loss.
\end{remark}

\begin{figure}[tbp]
\centering
\includegraphics[width=0.94\textwidth]{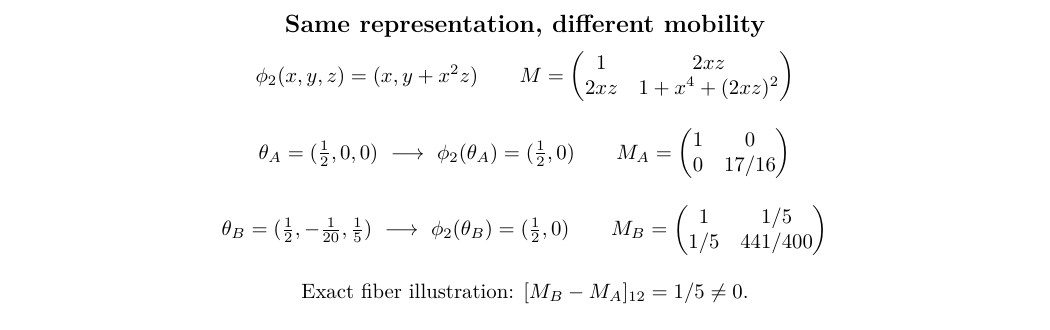}
\caption{An algebraic illustration for $\phi_2(x,y,z)=(x,y+x^2z)$.
Two states have the same representation and different mobility. These are
exact rational states on one fiber; this panel does not assert that they are
endpoints of two losses from the same initialization. The actual fixed-start
witnesses are proved in Proposition~\ref{ctrl:delayed} and
Theorem~\ref{loss:quadratic}. No integration or learning benchmark is shown.}
\label{fig:mobility}
\end{figure}
\section{Exact autonomous-loss exclusion controls}
The following constructions illustrate the logical strength of the ordinary
premise. They are algebraic witnesses, not experiments or separate novelty
claims. They do not refute the bridge theorem, because they fail its ordinary
closure assumption.

\begin{proposition}[Arbitrarily delayed brackets and a two-loss witness]
\label{ctrl:delayed}
For each integer $k\ge2$, let
$\phi_k(x,y,z)=(x,y+x^kz)$. Its representation rank is two and its gradient
Lie rank is three everywhere. At the origin all iterated brackets of lengths two through
$k$ vanish, while a bracket of length $k+1$ supplies the third direction.
The first differential mobility test passes at the origin, yet ordinary
closure fails there, as witnessed by two global smooth autonomous losses.
\end{proposition}
\noindent\textit{Proof.} See Appendix~\ref{release:proof28}.\par

\begin{proposition}[A rank-four affine-output exclusion]
For $\phi(x,y,z,w)=(x,y+xz+zw)$, $q=2$ and $p=4$ everywhere, and
ordinary closure fails at every initialization.
\label{release:unlabelled1}
\end{proposition}
\noindent\textit{Proof.} See Appendix~\ref{release:proof29}.\par

\section{What restricted loss families test}
\label{loss:section}

The arbitrary-loss quantifier in property \textup{(O)} is substantive.
For a family $\mathcal F$ of global output losses, write
$\textup{(O)}_{\mathcal F}$ for the same property with only $f\in\mathcal F$
tested: one parameter neighborhood $U$ and one entire matrix map
$K:\R^d\to\R^{d\times d}$, independent of $f$, must satisfy
\[
 M(\theta_f(t))=K(\phi(\theta_f(t)))
\]
on every forward local solution interval before exit from $U$, from the
same fixed initialization. The matrix $K$ is initially arbitrary.
In particular the equality concerns the entire mobility, rather than only
its product with the tested loss gradient. Property \textup{(O)} restricts
to every smaller family, but its necessity conclusions cannot be transferred
in the opposite direction without proof. The following supporting results
make this distinction explicit.

\begin{proposition}[Linear tests at a submersion]
\label{loss:linear}
Suppose $\phi$ is smooth and $D\phi(\theta_0)$ has full row rank $d$.
Then $\textup{(O)}_{\mathcal F_{\rm lin}}$ holds for
$\mathcal F_{\rm lin}=\{z\mapsto g^Tz:g\in\R^d\}$.
A common $K$ can be chosen globally smooth after shrinking the parameter
neighborhood. This holds for all $g$, without a common bound on its norm.
\end{proposition}
\noindent\textit{Proof.} See Appendix~\ref{release:proof30}.\par

\begin{remark}[Linear tests do not recover invariant-level factorization]
\label{loss:linear-separation}
For $k\ge2$, consider
\[
 \phi_k(x,y,z)=(x,y+x^kz),\qquad
 X=\partial_x,\quad Y=(kx^{k-1}z,1,x^k).
\]
The representation rank is two everywhere, and
$\operatorname{ad}_X^kY=k!\partial_z$ gives gradient Lie rank three.
Hence Proposition~\ref{loss:linear} applies. Nevertheless
property \textup{(I)} fails at the origin, directly at the level of its
actual conservation-law constraint. A common $\mathcal C^1$ law $h$
has $h_x=0$ and
\[
 h_y+x^kh_z=0
\]
on a product box. The first identity makes $h_y,h_z$ independent of $x$.
Taking $x=0$ and then any nonzero $x$ in the box forces
$h_y=h_z=0$. Thus every common law is locally constant, and any finite
common-law initial level contains a whole neighborhood. At fixed output
$(a,0)$ the parameter states $(a,-a^kz,z)$ have
$M_{12}=ka^{k-1}z$, which varies for arbitrarily small nonzero $a$ and $z$.
One represented mobility cannot serve that actual level.
\end{remark}

\begin{theorem}[One fixed strongly convex quadratic family]
\label{loss:quadratic}
Fix an integer $k\ge2$, a number $0<\gamma<1$, and
\[
 H=\begin{pmatrix}1&\gamma\\ \gamma&1\end{pmatrix},\qquad
 f_b(w)=\tfrac12w^THw-b^Tw,\quad b\in\R^2 .
\]
For $\phi_k(x,y,z)=(x,y+x^kz)$ initialized at the origin,
$\textup{(O)}_{\{f_b:b\in\R^2\}}$ fails. More precisely, for any fixed
$0<c_1<c_2$ there are $a_*>0$ and analytic parameter functions
$b_i(a)$ near $a=0$, with $b_i(0)=(1/c_i,0)$, such that the actual flows
satisfy, for every $0<a<a_*$,
\[
 \phi_k(\theta_i(c_i a))=(a,0),
\]
and
\begin{equation}\label{loss:gap}
 \bigl[M(\theta_2(c_2a))-M(\theta_1(c_1a))\bigr]_{12}
 =-\frac{\gamma k^2(c_2-c_1)}{2(k+1)(k+2)}
      a^{2k+1}+O(a^{2k+2})\ne0 .
\end{equation}
Both complete parameter paths converge uniformly to the origin as
$a\downarrow0$. The Hessian is fixed for both losses, with eigenvalues
$1-\gamma$ and $1+\gamma$; the translations and endpoint times differ.
\end{theorem}
\noindent\textit{Proof.} See Appendix~\ref{release:proof31}.\par

There is a second check on the coefficient in~\eqref{loss:gap}.
For the shooting trajectories constructed in Appendix~\ref{release:proof31},
fix $c\in\{c_1,c_2\}$ and write $s=t/(ca)$ and $x=aX(s)$.
Write $F(x)=x^k$, set $v=y+F(x)z$, and let $r_1,r_2$ be the negative
output-gradient components along any horizontal trajectory. Then
\[
 \dot x=r_1+F'zr_2,\quad
 \dot z=Fr_2,\quad
 \dot v=F'z\dot x+(1+F^2)r_2 .
\]
For a path with increasing $x$, elimination of time gives
\[
 \frac{d}{dx}\{z\sqrt{1+F^2}\}
   =\frac{F}{\sqrt{1+F^2}}v_x .
\]
Since the shooting endpoints have $v(0)=v(a)=0$, integration by
parts gives the exact hidden-coordinate identity
\[
 z(a)=-\frac{1}{\sqrt{1+F(a)^2}}
          \int_0^a\frac{v(x)F'(x)}{(1+F(x)^2)^{3/2}}\,dx .
\]
The shooting solution has $\dot x=1/c+O(a)>0$ uniformly. From the
limiting equations and smooth inversion of $s\mapsto X(s)$,
\[
 v(x)=\frac{c\gamma}{2}x(a-x)+O(a^3),\qquad 0\le x\le a .
\]
The uniform remainder contributes $O(a^{k+3})$ to $z(a)$;
denominator corrections contribute $O(a^{3k+2})$, which is absorbed
for $k\ge2$. The leading integral again gives
$-c\gamma k a^{k+2}/[2(k+1)(k+2)]$.

\begin{corollary}[Ordinary squared error with one fixed readout]
\label{loss:squared}
Let $W=H^{1/2}$ be the symmetric positive-definite square root of the
fixed Hessian in Theorem~\ref{loss:quadratic}, and fix the representation
$\psi_k=W\phi_k$. Its ordinary entire-mobility closure fails even when
the tested losses are only
\[
 \ell_\tau(w)=\tfrac12\|w-\tau\|_2^2,\qquad \tau\in\R^2.
\]
It has representation rank two and gradient Lie rank three everywhere,
and $DM_{\psi_k}(0)=0$. The compared targets and endpoint times differ;
the representation is fixed.
\end{corollary}
\noindent\textit{Proof.} See Appendix~\ref{release:proof32}.\par

\begin{remark}[Scope and a positive control]
The readout in Corollary~\ref{loss:squared} is a fixed change of represented
features, not a change in the Euclidean parameter metric.
It does not establish identity-Hessian failure on the original
untransformed $\phi_k$, a collision at equal times for the same fixed
Hessian, or a failure for one fixed target.
Choosing $\gamma=1/2$ fixes Hessian condition number three.
The corresponding projection representation $\psi_0=W(x,y)$ has
constant mobility $H$, and therefore admits closure for every loss.
Linear tests also pass for $\psi_k$ by Proposition~\ref{loss:linear}.
These statements distinguish the loss families; they do not constitute
a new optimizer, training gain or general quadratic-family classification.
\end{remark}

\begin{proposition}[Quadratic tests force section regularity]
\label{loss:regularity}
Suppose $\phi$ is smooth and has constant ambient rank $q>0$ on
a neighborhood of $\theta_0$. Set $z_0=\phi(\theta_0)$ and test only
\[
 f_b(z)=\tfrac12\|z-z_0\|_2^2+b^T(z-z_0),\qquad b\in\R^d .
\]
If $\textup{(O)}_{\{f_b:b\in\R^d\}}$ holds, then its arbitrary common $K$
agrees on a local represented image with a smooth representative.
That representative has a local ambient extension, and after shrinking
the image it has a global smooth extension.
If $\phi$ is analytic, the representative on the image and its local
extension can be analytic. This is an image-section conclusion, not
an invariant-level factorization.
\end{proposition}
\noindent\textit{Proof.} See Appendix~\ref{release:proof33}.\par

Proposition~\ref{loss:regularity} identifies $K$ on an actual endpoint
section. It says nothing by itself about other parameter points of the
same represented fiber. Proposition~\ref{loss:linear} and
Remark~\ref{loss:linear-separation} explain why section regularity alone
cannot substitute for the full arbitrary-loss necessity arguments.
The supporting results here remove a loss-family limitation of the
explicit negative control while retaining that distinction. They imply
no numerical value of $a_*$, empirical convergence order, universal
quadratic-family converse, or learning-performance claim.

\section{Related Work}\label{release:related}
\paragraph{Intrinsic dynamics and initialized tests.}
Marcotte, Peyr\'e, and Gribonval~\cite{intrinsic} introduce the ordinary
intrinsic-dynamic and invariant-metric notions used here, together with a
stronger definition for piecewise $\mathcal C^2$ autonomous losses.
Their Theorem~3.8 gives a positive relaxed-balance branch. Their Theorem~3.9
assumes full-rank factors and hidden width at most $\max(n,m)$, and excludes
their strong intrinsic dynamic property outside relaxed balance.
Proposition~3.10 already resolves $n=m=1$ at arbitrary width.
Our wide classification addresses ordinary closure with individually
full-row factors; its scalar subcase is inherited. Their Theorem~2.14 starts
from a differentiable invariant-level metric and Theorem~2.17 from recoverability.
The converse here starts from an unrestricted $K$ and fixed-start autonomous
tests, including specified cases with Lie rank larger than representation rank.

The conservation-law framework of Marcotte, Gribonval, and Peyr\'e~\cite{conservation},
Theorem~3.3 and Propositions~H.3/H.5, supplies the Lie distribution, the count
of independent common laws, and matrix factor-orbit ingredients. Their
Proposition~3.7 concerns recovery from invariants and representation.
The additional necessity step here passes from actual autonomous endpoint
patches to mobility constraints on Gram fibers. A factor-orbit dimension count
alone does not supply that step.

\paragraph{Known positive matrix flows and larger closed states.}
Domin\'e et al.~\cite{domine}, Theorem~5.2 and Appendix~C.2.1, recover the factors up to hidden orthogonal transformations, and thereby
determine the two mobility Gram blocks, under their $\lambda$-balanced and
full-rank-product assumptions, giving the square-root structure.
Bah et al.~\cite{bah}, Lemma~2 and Propositions~10--11, give the balanced
product flow and its Riemannian interpretation on fixed-rank matrix manifolds.
These are positive reduction results. The initialization exclusion and original
ambient invariant-level arguments in Theorem~\ref{mat:classification} answer
a different necessity question. Proposition~\ref{mat:knownpositive} explicitly
matches our positive flows to a known narrow family.

Zhang and Peyr\'e~\cite{zhang}, Proposition~2.1, close the flow on the full
weight covariance, retaining the hidden Gram blocks. The state here is only
$Z=UV^\top$. Importantly, their Proposition~E.2 also gives predictor-only
square-root closure on the hyperbolically isotropic leaf $CJC=\alpha^2J$
after the appropriate width normalization; this is the $a=b$ part of our
full-rank positive branch. For the unequal pair, the fixed shift
$D=C-(a-b)J/2$ satisfies $DJD=((a+b)/2)^2J$ and has the same covariance
flow because $H_GJ+JH_G=0$. Thus this positive reduction remains a known-family
consequence. Their Remark~E.3 shows that conserved spectral
values generally do not determine predictor velocity. That two-state example
does not itself establish ordinary failure from one fixed initialization.
We make no novelty claim for covariance closure or the positive square-root
formula, and do not infer a global priority claim from this comparison.

\paragraph{Geometric criteria and mirror flows.}
Li et al.~\cite{li}, Theorem~4.9, obtain mirror flow from a regular commuting
parametrization under their flow-domain assumptions. Theorem~4.10 gives a
necessary Lie-bracket condition when a mirror representation works for all
initializations and time-dependent losses. Mirror flow requires an integrable
inverse-Hessian metric, whereas properties (O) and (I) allow general mobility.
Our initialized autonomous-loss premise and invariant-level conclusion therefore
have different quantifiers and targets.
Horstmeyer and Atay~\cite{horstmeyer}, Proposition~1 and Theorem~4, characterize
exact lumpability by all-state projectability and invariance of the vertical
distribution under Lie differentiation. Their criterion applies to a fixed
vector field on smooth manifolds with the stated submersion/fiber hypotheses.
Applying it backwards here would first require proving nearby-fiber consistency
from initialized tests; this is precisely the extra implication we study.

Achour, Kohn, and Rauhut~\cite{achour}, Theorem~4 and Corollary~6 in the
April 2026 revision, show that for one-dimensional linear convolutional
networks with the specified strides larger than one, fixed norm-difference
invariants determine the NTK on the smooth neuromanifold, with finite fibers.
Their broader work also treats higher-dimensional convolutions. The finite-fiber
recovery geometry differs from the matrix-factorization fibers and $p>q$
regimes considered here. We do not extend the matrix classification to
convolutional architectures.

\paragraph{Optimization and implicit bias.}
Saxe et al.~\cite{saxe} provide exact deep-linear learning solutions for
structured settings. Arora et al.~\cite{arora18} interpret depth-induced
preconditioning; Gunasekar et al.~\cite{gunasekar} and Arora et al.~\cite{arora19}
study implicit regularization in matrix factorization. These motivate examining
represented dynamics, but a convergence or bias result under one loss is not
an all-loss mobility classification. Min et al.~\cite{min} bound convergence
using imbalance and product singular values. Nordio~\cite{nordio} studies
training-set dynamics with fixed readout and quadratic loss, retaining a
hierarchy of Gram operators at depth. Castin et al.~\cite{castin} use balancing
to change LoRA conditioning. Our results concern continuous Euclidean parameter
gradient flow and claim neither their discrete-time guarantees nor an improvement
in learning performance.

\section{Discussion and Limitations}
Theorem~\ref{mat:classification} determines when the prediction matrix can
serve as a complete mobility state for every autonomous loss from one fixed
initialization. In the nonscalar wide model, its generic nonclosure corollary
rules out that exact description under absolutely continuous initialization.
It leaves open approximate closure, loss-specific reductions, and dynamics on
larger states such as a full covariance.

The factors must be individually full row rank and $r>\max(n,m)$.
Individually deficient initializations are not classified. The general bridge
requires representation and gradient-Lie ranks constant on an ambient open
neighborhood; singular width reduction does not transfer that hypothesis.
The smooth case $q=2,p\ge4$ remains unresolved. All-order flatness alone
cannot replace equality of a smooth function on a neighborhood.
The fixed-Hessian quadratic witness is theoretical: no numerical $a_*$,
quadratic grid, or empirical convergence order is asserted. The historical
engineered-loss check in Appendix~\ref{num:sentinel} is a different test.
Continuous-time mobility equality does not imply equality of finite-step updates.

\paragraph{Use of AI-assisted tools.}
The research records document AI-assisted development of mathematical arguments,
source comparisons, code, and manuscript text, including automated critiques.
AI tools were also used to prepare this release, check algebra and references,
and inspect the build. These checks do not constitute verification by human
experts. The author is responsible for the mathematical
claims, citations, experimental reporting, and the decision to release the work.
No AI system is listed as an author.

\clearpage
\begingroup
\small
\bibliographystyle{unsrt}
\bibliography{references}
\endgroup
\clearpage
\appendix

\section{Proofs of the common local constructions}

\subsection{Proof of Lemma~\ref{lem:jet}}\label{release:proof1}
\begin{proof}
Choose a smooth orthonormal normal frame $e_a(t)$ along the extended interval.
In a tubular neighborhood write
$\Psi(t,y)=\gamma(t)+\sum_a y_a e_a(t)$ and define
\[
 A'(t)=\langle g(t),\dot\gamma(t)\rangle,\qquad
 f_{\rm loc}(\Psi(t,y))=A(t)+\sum_a y_a\langle g(t),e_a(t)\rangle.
\]
Tangential compatibility and the prescribed normal derivatives give
$\grad f_{\rm loc}(\gamma(t))=g(t)$. For $d=1$ there are no normal variables.
Multiply by a smooth cutoff supported inside the extended tube and identically
one near the original compact arc, and extend by zero to $\R^d$.
This is one global smooth loss. Uniqueness for its locally Lipschitz parameter
ODE identifies the prescribed path with the actual autonomous trajectory.
The slight extension prevents endpoint regularity problems. The scalar
compatibility also gives $A'=-g^TMg=-\|\dot\theta\|^2$ along the curve;
we have not prescribed inconsistent values around a projected loop.
\end{proof}

\subsection{Proof of Lemma~\ref{lem:clock}}\label{release:proof2}
\begin{proof}
Choose $g_c$ with $g_c^TM(\theta_0)g_c>0$. On a smaller state box and a bounded
convex open ball $B$ about $g_c$, continuity gives $g_c^TM(\theta)g\ge c>0$
for all $g\in B$. Thus along these controls
\[
 \frac{d}{dt}\langle g_c,\phi(\theta(t))\rangle
       =-g_c^TM(\theta(t))g(t)\le-c.
\]
Smooth controlled projected paths are regular and embedded. A compact state
box and uniform speed bound give a short horizon $\varepsilon$ keeping every
finite word of controls in $B$ with total duration less than $\varepsilon$
inside the chosen neighborhood. The bound depends on total time, not the
number of switches. Differences of members of $B$ span all controls, so their
Lie values still span $T L$.

Let $Q$ be the maximum differential rank of all positive-duration word endpoint
maps from $\theta_0$ with total time less than $\varepsilon$. Integer boundedness
ensures a word and a strictly positive duration vector attain $Q$. On a small
duration patch its rank is constantly $Q$, giving an embedded local image $P$.
Append a short positive segment of any fixed $Y_g$, $g\in B$. The augmented
endpoint map has rank at most $Q$ by maximality. Its $(Q+1)$-minors remain zero
as the appended duration decreases to zero, while the original $Q$-minor
persists. The appended derivative is $Y_g$, so $Y_g$ is tangent to $P$ throughout
a local patch. Brackets of tangent fields are tangent. The Lie spanning
hypothesis forces $Q\ge\dim L$, hence equality. This gives a word endpoint
image with nonempty interior in $L$, without claiming that it contains
$\theta_0$ or arbitrary return fibers.

Insert between successive fixed controls smooth flat-ended convex ramps of
one fixed duration $\delta$, keeping the plateau durations as parameters.
On a compact neighborhood, uniform field and state-Jacobian bounds imply
for each ramp map $T_\delta$
\[
 \|T_\delta-\Id\|\le C\delta,
 \qquad \|DT_\delta-I\|\le e^{C\delta}-1.
\]
These follow from the flow integral and variational equations; no bound on
the control's time derivatives is needed. Finite composition yields convergence
of the endpoint maps in state and duration derivatives. A full-rank minor
therefore persists for one sufficiently small positive $\delta$, with the
extra times within the strict horizon margin. The smoothed family itself,
not merely its limit, has an $L$-open endpoint image.
The clock inequality still holds and each control is constant near the outer
endpoints, so slight extensions remain embedded. Lemma~\ref{lem:jet} realizes
each by one global autonomous smooth loss. The common $K$ in \textup{(O)}
then gives the identity on the endpoint patch. No derivative or limit of an
unrestricted $K$ has been used.
\end{proof}

\subsection{Proof of Lemma~\ref{lem:section}}\label{release:proof3}
\begin{proof}
Constant rank gives a single local embedded image $S$ of dimension $q$.
Fix a sufficiently short $\tau>0$. For the global linear loss $f_g(z)=g^Tz$,
let $\theta_g(\tau)$ be its actual endpoint from $\theta_0$. Near $g=0$ all
paths remain in the ordinary neighborhood, and smooth dependence gives
\[
 D_g\theta_g(\tau)|_{g=0}=-\tau D\phi(\theta_0)^T,
 \qquad D_g\phi(\theta_g(\tau))|_{g=0}=-\tau M(\theta_0).
\]
Restricted to the $q$-dimensional space $\operatorname{im}D\phi(\theta_0)$,
the second derivative is an isomorphism onto $T_{z_0}S$. The inverse function
theorem gives a smooth endpoint section $s:S_0\to\Omega$ with
$\phi(s(z))=z$ on an image neighborhood $S_0$. Property \textup{(O)} forces
$K(z)=M(s(z))$ there. Flatten $S$ in target coordinates and extend this map
constantly in the normal coordinates to obtain $K_{\rm loc}$. Shrink the
parameter neighborhood so its whole image lies in $S_0$. Analytic dependence
of the linear-loss flow and the analytic inverse function theorem give the
analytic version. Off-image values of the original $K$ are irrelevant.
\end{proof}

\subsection{Proof of Lemma~\ref{lem:descent}}\label{release:proof4}
\begin{proof}
In a constant-rank product chart on $L$, the representation depends only on
$q$ coordinates and the assumed derivative identity makes $M$ independent
of the other coordinates. Restrict to a product with connected fibers to
factor $M$ through the local embedded image. Extend normally in target
coordinates to an open $W\ni z_0$, obtaining $K_{\rm loc}$. Choose a smooth
cutoff $\chi$ compactly supported in $W$, equal to one near $z_0$, and define
globally
\[
 \overline K=\chi K_{\rm loc}+(1-\chi)I_d,
\]
with the first term extended by zero off its support. Shrink the Frobenius
product box so its entire initial level lies in the factorization patch and
its represented image lies where $\chi=1$. This gives the actual local-level
identity and global smooth target extension. The argument is local and does
not identify disconnected global fibers. The analytic local descent follows
in analytic charts; the final cutoff is smooth.
\end{proof}

\section{Proofs of the wide matrix classification}

\subsection{Proof of Lemma~\ref{mat:factororbit}}\label{release:proof5}
\begin{proof}
The boosts \(H_{E_{ij}}\), together with their commutators, span all
matrices
\[
 T=\begin{bmatrix}P&G\\G^\top&Q\end{bmatrix},
       \qquad P^\top=-P,\quad Q^\top=-Q.
\]
Indeed, holding the bottom index fixed in two boosts produces, up to
sign, each elementary skew matrix in the top block; holding the top
index fixed produces each elementary skew matrix in the bottom block.
A block of dimension one requires no skew generators.

For the present $J$-orthogonal linear action, we use the classical
local orbit theorem on a homogeneous orbit restricted to an embedded
local orbit chart, with tangent
\(\{TF:T\in\mathfrak o(n,m)\}\), and the short flows of a
Lie-generating family generate its connected local orbit.
Only a local orbit chart is needed, not global closedness of the orbit.
For completeness its tangent rank is computed directly.
The map \(T\mapsto JT\) identifies \(\mathfrak o(n,m)\) with the
skew-symmetric matrices.  If \(W=\operatorname{im}F\), the condition
\(TF=0\) says that the corresponding alternating form annihilates \(W\)
in either argument.  These are exactly the alternating forms on
\(\R^N/W\); their dimension is \(\binom{N-s}2\).
Subtracting from \(\dim\mathfrak o(n,m)=\binom N2\) proves
\eqref{mat:factordim}.  Its rank is constant along the orbit, since
left multiplication and conjugation by a group element are invertible.

A common \(C^1\) conservation law has zero derivative along each
\(X_G\), so it is constant along every short boost flow, in either
time direction.  The local orbit statement therefore makes it
constant on the connected local orbit.  Alternatively one may use
finite boost products and their commutator limits, passing invariance
through continuity.  No second derivative of the conservation law is
taken.
\end{proof}

\subsection{Proof of Lemma~\ref{mat:gramobstruction}}\label{release:proof6}
\begin{proof}
Write \(\mathcal L\) for the connected local factor orbit.
It is a locally embedded invariant leaf and the generated Lie values
span its tangent space by Lemma~\ref{mat:factororbit}.
This is a statement on the orbit, even if ambient Lie rank jumps at
\(F_0\).  Moreover
\[
 \langle G,M_F[G]\rangle
      =\|G^\top U\|_{\mathrm F}^2+\|GV\|_{\mathrm F}^2.
\]
For \(F_0\ne0\) this positive semidefinite operator is nonzero.
Lemma~\ref{lem:clock}, applied on this orbit, supplies a nonempty
\(\mathcal L\)-open patch of endpoints of actual single global smooth
represented losses, on which
\[
                     M_F=K(\phi(F)).
\]
Its hypotheses require regularity along this invariant leaf, not
constant ambient rank at the singular factor.

Choose a point in that patch where the rank of
\(D(\phi|_{\mathcal L})\) is maximal within the patch.
A nonzero maximal minor persists, so the rank is constant on a smaller
open patch.  The constant-rank theorem supplies actual local fiber
curves there.  Equality through the same, possibly irregular \(K\)
makes \(M\) constant on these curves.  Consequently every orbit tangent
with \(\delta Z=0\) has \(\delta M=0\).
This differentiates the smooth \(M\) along a fiber, not the irregular
function \(K\).

The Gram map \(F\mapsto FF^\top\) intertwines the factor action and
the congruence action.  Every Gram-orbit tangent $TC+CT^\top$ lifts to the factor tangent $TF$.  Its differential rank is constant along the
factor orbit by invertibility of both actions; its tangent image is
the Gram-orbit tangent.  Lift any such tangent
\[
 \delta C=\begin{bmatrix}\delta A&0\\0&\delta B\end{bmatrix}
\]
to a factor-orbit tangent.  The preceding fiber statement gives
\(\delta A G+G\delta B=0\) for every \(G\).
Testing matrix units first eliminates all off-diagonal entries and
then gives
\[
                    \delta A=t\Id_n,\qquad
                    \delta B=-t\Id_m
\]
for one scalar \(t\).
The congruence orbit preserves \(\tr(JC)\), because \(Q^\top JQ=J\)
implies \(\tr(JQCQ^\top)=\tr(JC)\).  Therefore
\(0=\tr(J\delta C)=Nt\), so \(\delta C=0\).
The Gram tangent projects injectively to the \(nm\)-dimensional
\(Z\)-space at the chosen point.  Its dimension is constant on the
orbit, proving \(d_C\leq nm\) at \(C_0\).

If \(\textup{(I)}\) holds, it implies \(\textup{(O)}\) as noted above.
One can also work directly on the local orbit, where every chosen
common conservation law is constant, and differentiate the \(C^1\)
represented factorization.  Thus no constant-rank hypothesis on \(Dh\)
is needed.
\end{proof}

\subsection{Proof of Lemma~\ref{mat:column}}\label{release:proof7}
\begin{proof}
The differential of \(F\mapsto S\) is onto the symmetric matrices.
For any symmetric \(E\), the choice
\[
             \delta F=\tfrac12 JF(F^\top F)^{-1}E
\]
gives \(F^\top J\delta F=E/2\) and \(\delta S=E\).
Thus its kernel has dimension \(Ns-s(s+1)/2\), exactly
\eqref{mat:factordim}.  Every factor-orbit tangent lies in this kernel,
so the two tangent spaces coincide.

The kernel of the differential of \(F\mapsto FF^\top\) is
\(\delta F=F\Omega\) with \(\Omega^\top=-\Omega\).
To prove the nontrivial direction, project \(\delta C=0\) onto
\(W^\perp\), where \(W=\operatorname{im}F\), and multiply on the right
by \(F(F^\top F)^{-1}\).  This gives
\(P_{W^\perp}\delta F=0\), so \(\delta F=F\Omega\).
Then \(F(\Omega+\Omega^\top)F^\top=0\) and a left inverse of \(F\)
makes \(\Omega\) skew.

For these variations,
\[
             \delta S=\Omega^\top S+S\Omega
                         =S\Omega-\Omega S.
\]
The Gram differential kernel restricted to the orbit is therefore
exactly the skew matrices commuting with \(S\).
An orthogonal spectral decomposition of this real symmetric matrix
makes that kernel the direct sum of skew matrices on its eigenspaces,
of dimension \(\sum_j\binom{c_j}2\).  Subtraction proves
\eqref{mat:gramcolumn}, with the null eigenspace included.
\end{proof}

\subsection{Proof of Lemma~\ref{mat:intermediate}}\label{release:proof8}
\begin{proof}
Let \((\alpha,\beta,\kappa)\) be the inertia of \(S\).
Thus \(\alpha+\beta+\kappa=s\).
By Sylvester's law of inertia, remove a nondegenerate subspace of
signature \((\alpha,\beta)\) from \(\operatorname{im}F\).
Its \(J\)-orthogonal complement has signature
\((n-\alpha,m-\beta)\) and contains the \(\kappa\)-dimensional
totally isotropic radical.  Projection of a totally isotropic subspace
into either definite summand is injective.  Hence
\(\kappa\leq\min(n-\alpha,m-\beta)\).
Put
\[
 u=n-\alpha-\kappa\geq0,\qquad
 v=m-\beta-\kappa\geq0.
\]
The strict width bounds give \(\beta>u\), \(\alpha>v\), and
\(N-s=\kappa+u+v>0\).  In particular \(\alpha,\beta>0\).
Equal eigenvalues cannot belong to different signs.  Grouping each
sign, and the zero cluster, into a single cluster maximizes
the possible centralizer dimension:
\[
       \sum_j\binom{c_j}2
         \leq\binom{\alpha}2+\binom{\beta}2+\binom{\kappa}2.
\]
Substitution into \eqref{mat:gramcolumn}, with
\(n=\alpha+\kappa+u\) and \(m=\beta+\kappa+v\), gives
\[
 \begin{aligned}
 d_C-nm
 &\geq \alpha u+\beta v+\kappa(\alpha+\beta)-uv\\
 &=u(\alpha-v)+\beta v+\kappa(\alpha+\beta)>0.
 \end{aligned}
\]
If \(u>0\), the first term is positive; if \(v>0\), the second is
positive.  If both vanish, \(\kappa>0\) and the last term is positive.
These cover every case, including a nonzero radical.
\end{proof}

\subsection{Proof of Lemma~\ref{mat:minimal}}\label{release:proof9}
\begin{proof}
Since \(s=L_0\), direct substitution into \eqref{mat:gramcolumn} gives
\[
 d_C-nm=\binom s2-\sum_j\binom{c_j}2
                  =\sum_{i<j}c_ic_j.
\]
It is zero precisely when \(S\) has one eigenvalue value.
As \(CJC=FSF^\top\), scalar \(S\) implies \(CJC=\lambda C\);
a left inverse of \(F\) proves the converse.

If \(n>m\), then \(s=n\), the reduced \(U\) is square invertible,
and a nonzero \(x\in\ker V\) satisfies
\(\lambda\|x\|^2=\|Ux\|^2>0\).
The case \(m>n\) is symmetric and yields \(\lambda<0\).
If \(n=m=s\), both blocks are square invertible and
\(Z=UV^\top\) is invertible, without a required sign of \(\lambda\).
\end{proof}

\subsection{Proof of Lemma~\ref{mat:row}}\label{release:proof10}
\begin{proof}
The symmetric matrix \(L=C^{1/2}JC^{1/2}\) has inertia \((n,m)\)
and is similar to \(JC\).
Choose an orthogonal \(O\) such that
\[
 O^\top LO=\diag(a_1,\ldots,a_n,-b_1,\ldots,-b_m),
                 \qquad a_i,b_j>0,
\]
and put \(D=\diag(a_1,\ldots,a_n,b_1,\ldots,b_m)\).
The matrix \(R=C^{1/2}OD^{-1/2}\) satisfies
\[
                         R^\top JR=J,\qquad RDR^\top=C.
\]
We may therefore compute the congruence stabilizer at \(D\).
For \(T=\left[\begin{smallmatrix}P&G\\G^\top&Q\end{smallmatrix}\right]\)
in \(\mathfrak o(n,m)\), the equation \(TD+DT^\top=0\) requires
\[
 [P,\diag(a_i)]=0,\quad [Q,\diag(b_j)]=0,\quad
              \diag(a_i)G+G\diag(b_j)=0.
\]
The last equation forces \(G=0\), entry by entry.
The stabilizer dimension is
\(\sum_i\binom{c_i}2+\sum_j\binom{d_j}2\).
The classical orbit--stabilizer dimension formula for this smooth
finite-dimensional action now gives
\[
          d_C=\binom N2-\sum_i\binom{c_i}2-\sum_j\binom{d_j}2,
\]
which is \eqref{mat:gramrow}.
Each numerator \(n^2-\sum_i c_i^2\) or \(m^2-\sum_jd_j^2\)
is nonnegative, and vanishes exactly when that sign has one cluster.

We next establish sufficiency on the actual ambient invariant level.
The identity
\[
            H_GJ+JH_G=0
\]
shows by differentiation of \eqref{mat:factorflow} that
\(S=F^\top JF\) is common to every represented-loss flow.
For a full-row \(F_0\), \(S_0\) has rank \(N\).
If \(F^\top JF=S_0\), then \(F\) also has full row rank, and
\[
       \operatorname{im}F^\top
          =\operatorname{im}S_0=\operatorname{im}F_0^\top.
\]
For example, \(\operatorname{im}(F^\top JF)=
\operatorname{im}F^\top\) follows because \(F\) is onto and \(J\)
is invertible.  Equality of row spaces yields \(F=QF_0\) for an
invertible \(Q\); multiplication by a right inverse of \(F_0\) and its
transpose makes the equality of canonical invariants equivalent to
\(Q^\top JQ=J\).  Conversely every such \(Q\) preserves that invariant.
Thus the canonical level is precisely the \(J\)-orthogonal factor
orbit, not only its tangent space.

Since \(JC_0\) is diagonalizable with the two stated values, it obeys
\((JC_0-a\Id)(JC_0+b\Id)=0\).  Multiplication by \(J\) gives
\begin{equation}
       C_0JC_0-(a-b)C_0-abJ=0. \label{mat:spectralidentity}
\end{equation}
This identity is preserved by \(J\)-orthogonal congruence, using also
\(QJQ^\top=J\).  Put \(c=a-b\) and \(\sigma=(a+b)/2\).
On the whole canonical level its diagonal blocks give
\[
       A^2-cA=ZZ^\top+ab\Id_n,\qquad
       B^2+cB=Z^\top Z+ab\Id_m.
\]
The block Grams are positive definite.  Diagonalizing either Gram
also diagonalizes the polynomial on the other side.
Each scalar quadratic has exactly one positive root, because the
product of its roots is negative.  Since
\(ab+c^2/4=\sigma^2\), this yields
\[
 A=c\Id_n/2+\sqrt{\sigma^2\Id_n+ZZ^\top},\qquad
 B=-c\Id_m/2+\sqrt{\sigma^2\Id_m+Z^\top Z}.
\]
The scalar shifts cancel in \eqref{mat:mobility}.
The positive definite square-root map is smooth, and
\(\sigma>0\) makes the displayed represented formula smooth for every
\(Z\).  The finite entries of \(S\) and this globally smooth map
therefore prove \(\textup{(I)}\).
\end{proof}

\subsection{Proof of Lemma~\ref{mat:reduction}}\label{release:proof11}
\begin{proof}
The right action \(F\mapsto FQ\) is a Euclidean isometry preserving
\(UV^\top\) and \(FF^\top\), hence intertwines the gradient flows.
Along any actual trajectory, \eqref{mat:factorflow} is a linear
left equation with continuous coefficient \(-H_{\grad f(Z(t))}\).
Its fundamental solution \(P(t)\) is invertible at every finite time
in the existence interval, so \(F(t)=P(t)F_0\) and
\(\ker F(t)=\ker F_0\).
The zero hidden columns remain zero and the other columns satisfy
exactly the reduced system.  Their Gram quantities and represented
variables coincide.

If \(\textup{(O)}\) holds in the original neighborhood, pull that
neighborhood back under the continuous embedding
\(\widetilde F\mapsto[\widetilde F,0]Q^\top\).
If it holds on a reduced neighborhood, take its inverse image under
the projection onto the first \(s\) columns of \(FQ\).
This is an open original neighborhood, and the actual original
trajectory always has zero remaining columns.  The same \(K\) works
in each direction for every loss.
These arguments concern trajectories from the fixed initialization;
they make no assertion about arbitrary ambient points on an invariant
level.  Finally an orthogonal column action and deletion of identically
zero columns preserve each block's row rank.
\end{proof}

\subsection{Proof of Lemma~\ref{mat:ambientboundary}}\label{release:proof12}
\begin{proof}
Use the fixed hidden orthogonal coordinates of
Lemma~\ref{mat:reduction}.
Lemma~\ref{mat:minimal} gives
\[
                 S_0=\diag(\lambda\Id_s,0).
\]
Consider an arbitrary ambient \(F=[W,E]\) on this exact canonical
level.  It satisfies
\begin{equation}
 W^\top JW=\lambda\Id_s,\qquad
 W^\top JE=0,\qquad E^\top JE=0.
 \label{mat:levelblocks}
\end{equation}

First suppose \(\lambda\ne0\).
The sign restrictions in Lemma~\ref{mat:minimal} show that
\(\lambda>0\) entails \(s=n\), and \(\lambda<0\) entails \(s=m\);
this includes the square case \(n=m=s\).
If \(\lambda>0\), the subspace spanned by \(W\) is positive definite
of dimension \(n\), exhausting the positive index of \(J\).
Its \(J\)-orthogonal complement is negative definite.
Each column of \(E\) lies in that complement and has zero \(J\)-square
by \eqref{mat:levelblocks}, hence is zero.
For \(\lambda<0\) the complementary form is positive definite
and the same conclusion follows.
Thus every ambient point of this invariant level has \(E=0\)
and \(CJC=\lambda C\).

On a small neighborhood of \(F_0\), both block Grams remain positive
definite, by the assumed separate full row ranks.
The block equations of \(CJC=\lambda C\) are
\[
              A^2-\lambda A=ZZ^\top,\qquad
              B^2+\lambda B=Z^\top Z.
\]
For every \(\mu\geq0\), \(x^2-\lambda x-\mu=0\) has at most
one positive root; any existing positive Gram eigenvalue selects it.
The corresponding statement with \(-\lambda\) applies to \(B\).
Functional calculus therefore yields
\begin{equation}
 \begin{aligned}
 A&=\lambda\Id_n/2+\sqrt{\lambda^2\Id_n/4+ZZ^\top},\\
 B&=-\lambda\Id_m/2+\sqrt{\lambda^2\Id_m/4+Z^\top Z}.
 \end{aligned}
 \label{mat:boundaryroots}
\end{equation}
Their shifts cancel.  Since \(\lambda\ne0\), the resulting
\(K_{|\lambda|/2}\) is smooth on the entire represented space.

Now suppose \(\lambda=0\).
The sign restrictions force \(n=m=s\), and the full original
\(S_0\) is zero.  We do not discard or force the extra hidden
columns to vanish.  At every point of the actual canonical level,
\[
                   CJC=F(F^\top JF)F^\top=0.
\]
Consequently \(A^2=ZZ^\top\) and \(B^2=Z^\top Z\).
Both Grams remain positive definite on a sufficiently small ambient
neighborhood, so \(Z\) is invertible and their positive roots give
\eqref{mat:boundaryroots} with \(\lambda=0\).
The map \(K_0\) is smooth on \(\operatorname{GL}(n,\R)\).
Choose a smooth cutoff \(\chi\) compactly supported in that open set,
equal to one near \(Z_0\), and define globally
\[
 \overline K(Z)=\chi(Z)K_0(Z)+(1-\chi(Z))\Id_{\R^{n\times n}},
\]
where the first term is extended by zero off its compact support.
This is globally smooth.  Shrink the ambient parameter neighborhood
so its represented image lies in the region \(\chi=1\).
In both cases the finite canonical entries are common smooth
conservation laws, and the required identity holds on their actual
initial level.  This proves \(\textup{(I)}\) in the original space.
\end{proof}

\subsection{Proof of Theorem~\ref{mat:classification}}\label{release:proof13}
\begin{proof}[Proof of Theorem~\ref{mat:classification}]
Separate full row ranks imply \(L_0\leq s\leq\min(r,N)\).
Apply Lemma~\ref{mat:reduction} to property \(\textup{(O)}\).
If \(L_0<s<N\), the reduced factor is full column rank and
Lemmas~\ref{mat:gramobstruction} and \ref{mat:intermediate}
contradict each other.
If \(s=L_0\), Lemmas~\ref{mat:gramobstruction} and \ref{mat:minimal}
force \eqref{mat:mincase} and its signs.
If \(s=N\), the full-row calculation
\eqref{mat:gramrow} forces \eqref{mat:fullcase}.
These exhaust every admissible joint rank.

Conversely Lemma~\ref{mat:ambientboundary} proves
\(\textup{(I)}\) in \eqref{mat:mincase}, and
Lemma~\ref{mat:row} proves it in \eqref{mat:fullcase}.
Both use the actual original canonical level, not a reduced-space
tangent argument.  Conservation gives \(\textup{(O)}\).
The mobility formulas were established in the same lemmas.
\end{proof}

\subsection{Proof of Proposition~\ref{mat:knownpositive}}\label{release:proof14}
\begin{proof}
Put
\[
 P=\sqrt{\sigma^2\Id_n+ZZ^\top},\qquad
 Q=\sqrt{\sigma^2\Id_m+Z^\top Z}.
\]
If \(n\leq m\), define
\[
 U'=(P+\sigma\Id_n)^{1/2},\qquad
 V'=Z^\top(P+\sigma\Id_n)^{-1/2}.
\]
The inverse is lawful since \(P+\sigma\Id_n\geq2\sigma\Id_n\).
The product is \(U'V'^\top=Z\).
A singular-value decomposition, including all zero singular values,
gives
\[
 \begin{aligned}
 U'U'^\top&=P+\sigma\Id_n,&
 V'V'^\top&=Z^\top(P+\sigma\Id_n)^{-1}Z=Q-\sigma\Id_m,\\
 U'^\top U'&=P+\sigma\Id_n,&
 V'^\top V'&=P-\sigma\Id_n.
 \end{aligned}
\]
For a singular value \(d\), the scalar identity used here is
\[
 \frac{d^2}{\sqrt{\sigma^2+d^2}+\sigma}
                      =\sqrt{\sigma^2+d^2}-\sigma.
\]
It also handles the extra nullspace, where both sides are zero.
The hidden imbalance is \(2\sigma\Id_n\), and its two scalar Gram
shifts cancel in the mobility.

If \(n>m\), the transposed construction is
\[
 V'=(Q+\sigma\Id_m)^{1/2},\qquad
 U'=Z(Q+\sigma\Id_m)^{-1/2}.
\]
It gives
\[
 U'U'^\top=P-\sigma\Id_n,\quad
 V'V'^\top=Q+\sigma\Id_m,\quad
 U'^\top U'-V'^\top V'=-2\sigma\Id_m,
\]
and the same cancellation.

These scalar hidden invariants remain constant along each narrow
factor flow.  In the first construction \(U'\) stays invertible,
since \(U'^\top U'\geq2\sigma\Id_n\).
Its Gram equation has the unique positive solution
\(U'U'^\top=P+\sigma\Id_n\); the other block has the unique
nonnegative solution \(V'V'^\top=Q-\sigma\Id_m\).
The latter remains unique even at zero singular values, because
\(x^2+2\sigma x-\mu=0\), \(\mu\geq0\), has a unique nonnegative
root.  The transposed construction has the identical argument.
Hence the narrow flow retains exactly \(K_\sigma\).
For square invertible \(Z\) and \(\sigma=0\), set
\(U'=P^{1/2}\), \(V'=Z^\top P^{-1/2}\); its hidden imbalance is zero
and its Grams are \(P,Q\), giving the balanced formula locally.

For each common represented \(C^2\) loss, the wide positive flow and
these narrow factors therefore solve the same smooth represented ODE,
with the same initial represented matrix.
Local uniqueness identifies their represented trajectories on a
common existence interval.
This is a mobility and continuous-time trajectory identity, not
equality of factor states or finite-step gradient descent.
The scalar-balanced square-root dynamics are the known positive
family; the construction supplies the exact matching comparison.
\end{proof}

\subsection{Proof of Corollary~\ref{mat:generic}}\label{release:proof15}
\begin{proof}
Separate maximal row ranks and maximal joint rank hold almost surely:
their failures lie in zero sets of nonzero minor polynomials.
If \(r<N\), the almost-sure joint rank is \(r\), strictly between
\(L_0\) and \(N\), so Theorem~\ref{mat:classification} excludes closure.
If \(r\geq N\), the almost-sure joint rank is \(N\).
For \((n,m)\ne(1,1)\), the positive spectral case forces a repeated
eigenvalue of \(JFF^\top\).
The discriminant
\[
             P(F)=\operatorname{disc}_t
                         \det(t\Id_N-JFF^\top)
\]
is a polynomial in the factor entries and must vanish there.
It is not identically zero: at
\(F=[\diag(1,\ldots,N),0]\) its signed diagonal eigenvalues
are all distinct.
The zero set of a nonzero real polynomial has Lebesgue measure zero,
and hence probability zero under every such law.
For \(n=m=1\), the two factor rows are nonzero almost surely.
Joint rank one automatically gives \(CJC=\lambda C\);
joint rank two automatically gives one positive and one negative
signed spectral value.  The theorem then gives closure, in agreement with the known scalar case \cite[Proposition 3.10]{intrinsic}.
\end{proof}

\section{Proofs of the geometric characterization}

\subsection{Proof of Lemma~\ref{bridge:plaque}}\label{release:proof16}
\begin{proof}
Select finitely many bracket fields $V_1,\ldots,V_p$ independent at
$\theta_0$.  A nonzero frame minor stays nonzero on a smaller
neighborhood; constant rank makes these fields a frame for
$\mathcal E$ there.  Their brackets belong pointwise to $\mathcal E$,
and their frame coefficients are smooth, or analytic in the analytic
case, by inversion of the fixed frame minor.  Leibniz's rule then makes
this subbundle involutive.  The corresponding smooth or analytic
Frobenius theorem supplies coordinates $(x,h)$ in a product box, with
$\mathcal E=\ker Dh$.  The components of $h$ annihilate every
$\grad\phi_i$, so they are common conservation laws.  The set
$h=h(\theta_0)$ in this box is a single connected plaque.  If $p=D$,
one may take $h$ constant.

Because $\mathcal H\subset\mathcal E=T L$, restriction of $D\phi$ to
$T L$ has image containing
\[
 \operatorname{image}(D\phi D\phi^{\mathsf T})
   =\operatorname{image}D\phi.
\]
The reverse containment is automatic.  Thus the restriction has the
same rank $q$.  No full-output-rank assumption is involved.
\end{proof}

\subsection{Proof of Theorem~\ref{bridge:main} in the analytic case}\label{release:proof17}
\begin{proof}[Proof of Theorem~\ref{bridge:main} in the analytic case]
The metric property implies ordinary closure by conservation, so
consider the converse.  If $q=0$, $D\phi=0$ locally, $M=0$, and the
zero global mobility and a constant conservation law suffice.  Assume
$q>0$.  Use the connected analytic plaque $L$ of
Lemma~\ref{bridge:plaque}.  The clock endpoint construction in
Lemma~\ref{lem:clock} gives a nonempty relatively open set $O\subset L$
on which
\[
 M(\theta)=K(\phi(\theta)),
\]
with the same initially arbitrary $K$ as in ordinary closure.  Every
vector in $\ker D(\phi|_L)$ at a point of $O$ is tangent to an actual
local fiber curve, because the restriction has constant rank $q$.
Along such a curve $\phi$ is constant, hence so is $M$.  Therefore
\begin{equation}
 \ker D(\phi|_L)\subset\ker D(M|_L)\quad\hbox{on }O.
 \label{bridge:vertical-condition}
\end{equation}
This argument neither differentiates $K$ nor assumes its continuity.

Represent $M$ by its $d^2$ entries.  The augmented differential
$D_L(\phi,M)$ has rank $q$ on $O$.  In the connected analytic
$x$-coordinate box of $L$, every $(q+1)$-minor of this differential is
analytic and vanishes on $O$.  The real-analytic identity principle
makes every such minor zero throughout that box.  If $p=q$, the upper
bound is automatic since there are only $p$ columns.  Since
$D(\phi|_L)$ has rank $q$ everywhere, the augmented differential has
rank exactly $q$ everywhere.  Thus the kernel inclusion above holds on
the entire plaque, including near $\theta_0$.

For completeness, this differential condition gives an equality on the
actual local level, not just on its tangent.  Apply the analytic
constant-rank theorem to $\phi|_L$ near $\theta_0$.  In local source
coordinates $(s,t)\in\R^q\times\R^{p-q}$ it has the form
$\phi(s,t)=\iota(s)$, where $\iota$ parametrizes a locally embedded
analytic image $S$.  Shrink to a product box so its $t$-fibers are
connected.  The kernel inclusion says $\partial_tM=0$, hence
$M(s,t)=m(s)$.  Taking a local section, for example $t=0$, constructs
an analytic map $k$ on $S$ with $M=k\circ\phi$ on this whole smaller
plaque neighborhood.

Flatten $S$ in analytic ambient target coordinates and extend $k$
constantly in the normal coordinates, obtaining $K_{\mathrm{loc}}$ on
an open target neighborhood $W$ of $\phi(\theta_0)$.  Choose a smooth
$0\leq\chi\leq1$ with compact support in $W$ and equal to one on a
smaller neighborhood.  The map
\[
 \overline K(z)=\chi(z)K_{\mathrm{loc}}(z)+(1-\chi(z))\Id_d
\]
is globally smooth, interpreting $\chi K_{\mathrm{loc}}$ as zero
outside $W$.  No analyticity of the global cutoff is asserted.
Finally shrink the original Frobenius product box so its entire
initial $h$-level lies in the factorization neighborhood and its
represented image lies where $\chi=1$.  This proves the exact local
metric property.  The $h$ are local conservation laws; they need not
extend to a larger originally specified parameter domain.
\end{proof}

\subsection{Proof of Lemma~\ref{bridge:embedded}}\label{release:proof18}
\begin{proof}
A self-intersection at $s\ne t$ would require
\[
 a=-\frac{w(t)-w(s)}{t-s}.
\]
Cover the off-diagonal square by countably many compact
two-dimensional pieces bounded away from the diagonal.  The secant
map is smooth and Lipschitz on each piece.  Its image has
$q$-dimensional Lebesgue measure zero when $q>2$, by the elementary
covering estimate for a Lipschitz image of a two-dimensional set.
Failure of immersion requires $a=-w'(t)$; this one-dimensional smooth
image is also null.  Choose arbitrarily small $a_j$ outside their
union.  Each resulting curve is injective and regular; compactness
makes it an embedded compact arc.  The $C^1$ convergence is immediate,
and compact containment gives the last claim.  Nothing in this proof
asserts density of embedded planar arcs when $q=2$.
\end{proof}

\subsection{Proof of Lemma~\ref{bridge:horizontal}}\label{release:proof19}
\begin{proof}
Construct $K_{\mathrm{loc}}$ first, and shrink into its domain and the
ordinary neighborhood.  Constant rank gives one embedded represented
image $S$ and local coordinates $\psi:S\to\R^q$.  Set
$\pi=\psi\circ\phi$.  The kernels of $D\pi$ and $D\phi$ agree, so
their Euclidean horizontal spaces agree.  The smooth right inverse
\[
 H(\theta)=D\pi(\theta)^{\mathsf T}
    [D\pi(\theta)D\pi(\theta)^{\mathsf T}]^{-1}
\]
maps base velocities to their unique horizontal lifts.

For a smooth horizontal path $\theta(t)$, put
$w(t)=\pi(\theta(t))$.  It solves $\theta'=H(\theta)w'$.  Apply
Lemma~\ref{bridge:embedded} and lift $w_j$ from exactly $\theta_0$ by
$\theta_j'=H(\theta_j)w_j'$.  Differentiation gives
$\pi(\theta_j(t))=w_j(t)$.  Standard ODE stability on a compact state
neighborhood gives uniform convergence of lifts on $[0,T]$.  One
explicit bound, while both lifts remain there, is
\[
 \sup_{t\leq T}\|\theta_j(t)-\theta(t)\|
 \leq C\|w_j'-w'\|_{L^1}
       \exp\!\left(C\int_0^T\|w_j'(t)\|\,dt\right).
\]
A slightly larger compact neighborhood and continuation show that all
large-$j$ lifts exist on the full interval and remain in the required
open set.  Their represented arcs $z_j=\phi(\theta_j)=\psi^{-1}(w_j)$
are embedded and regular.

Here is the exact autonomous realization used for each arc.  The
constant-rank Moore--Penrose inverse $M^\dagger$ is smooth.  Put
$g_j=-M(\theta_j)^\dagger z_j'$.  Horizontality gives
$-D\phi^{\mathsf T}g_j=\theta_j'$ and
\[
 \langle g_j,z_j'\rangle=-\|\theta_j'\|^2<0.
\]
Slightly extend the regular embedded arc at its two ends.  In tubular
coordinates $z_j(t)+\sum_b y_b e_b(t)$, where the $e_b$ form a smooth
normal frame, prescribe
\[
 A'(t)=\langle g_j(t),z_j'(t)\rangle,
 \qquad f_{\mathrm{loc}}(t,y)
 =A(t)+\sum_b y_b\langle g_j(t),e_b(t)\rangle.
\]
Its gradient on the arc is exactly $g_j$: the primitive supplies the
tangential derivative, and the linear terms supply all normal
derivatives.  A cutoff equal to one around the original compact arc
and supported in this slightly extended tube yields one global smooth
represented loss.  ODE uniqueness identifies its actual parameter
flow from $\theta_0$ with $\theta_j$.  The construction requires no
uniform derivative bound over $j$.

Ordinary closure therefore gives
$M(\theta_j(t))=K(\phi(\theta_j(t)))
=K_{\mathrm{loc}}(\phi(\theta_j(t)))$.
Pass to the limit through the separately forced smooth
$K_{\mathrm{loc}}$.  This proves the asserted identity on the original
horizontal path without assuming continuity of the original $K$.
\end{proof}

\subsection{Proof of Lemma~\ref{bridge:accessibility}}\label{release:proof20}
\begin{proof}
Use positive durations and the finite symmetric field family
$\{\grad\phi_i,-\grad\phi_i\}$.  A uniform speed bound on a compact
chart gives a sufficiently short total horizon for all words under
consideration.  Maximize the differential rank of their endpoint maps
over all finite words and strictly positive duration vectors.  The
maximum $Q$ is attained, since ranks are integers bounded by $p$.
Around a maximizing vector, a nonzero $Q$-minor persists and no higher
rank is possible.  The constant-rank theorem gives an embedded image
patch $P$ of dimension $Q$ with a strict remaining horizon margin.

Append any one of the allowed fields for an additional positive
duration.  Maximality bounds the augmented rank by $Q$.  Taking that
duration to zero makes its extra derivative tangent to $P$; the
original $Q$-minor still persists.  Thus all the generators, and hence
their brackets, are tangent to $P$.  Their Lie rank is $p$, so $Q=p$
and $P$ is open in $L$.

Let the fixed maximizing word end at $\theta_*$.  Its inverse word is
a local diffeomorphism sending $\theta_*$ to $\theta_0$ and a smaller
open patch $P$ to a neighborhood of $\theta_0$ in $L$.  Every point of
that neighborhood is obtained by a nearby forward word followed by
this fixed inverse word.  Finite-path continuous dependence ensures
that all these paths stay in the larger chart after shrinking $P$.
This uses control words only as horizontal paths; it does not call
their concatenation a single autonomous loss.

For each segment, replace its time parameter by a smooth increasing
function with the same endpoints and all derivatives zero at both
ends.  The resulting joined path is smooth and horizontal, including
at the joins, and has exactly the same image and endpoint.
\end{proof}

\subsection{\texorpdfstring{Proof of Theorem~\ref{bridge:main} for $q\geq3$ and the elementary low-rank cases}{Proof in the specified smooth regime}}\label{release:proof21}
\begin{proof}[Proof of Theorem~\ref{bridge:main} for $q\geq3$ and the elementary low-rank cases]
For $q\geq3$, Lemmas~\ref{bridge:horizontal} and
\ref{bridge:accessibility} give the mobility identity on an entire
neighborhood of $\theta_0$ in its initial plaque.  Shrink the original
Frobenius product box so its actual initial level lies there.  A target
cutoff as in the analytic proof makes $K_{\mathrm{loc}}$ globally
smooth and proves the metric property.

If $q=0$, the stationary argument already applies.  If $q=1$, the
regular horizontal line bundle is involutive, so $p=1$.  More
generally, whenever $p=q$, $\phi$ is locally injective on the initial
plaque by the constant-rank theorem.  A local section factors $M$
smoothly through its represented image, and target extension and
cutoff give the metric property without assuming ordinary closure.
These are the classical local recoverability cases.  It remains to
prove the smooth $q=2,p=3$ case.
\end{proof}

\subsection{Proof of Lemma~\ref{bridge:curvature-jet}}\label{release:proof22}
\begin{proof}
The coordinate-gradient fields and $H_1,H_2$ are smooth frame changes
of the same horizontal bundle.  Their generated Lie modules have the
same evaluated distribution, namely $T L$.

Let $\mathcal I$ be the smooth ideal generated by all ordered
$H_i$-derivatives of $c$, including $c$.  It is stable under $H_1,H_2$
by Leibniz's rule.  The module
\[
 \mathcal N=\Span_{C^\infty}\{H_1,H_2\}
                +\mathcal I\partial_\eta
\]
is closed under brackets.  In fact, with $A_1=A,A_2=B$,
\[
 [H_i,a\partial_\eta]=(H_i a-a\partial_\eta A_i)\partial_\eta,
 \quad
 [a\partial_\eta,b\partial_\eta]
       =(a\partial_\eta b-b\partial_\eta a)\partial_\eta.
\]
Both coefficients belong to $\mathcal I$ when $a,b\in\mathcal I$;
stability under $\partial_\eta$ is not needed.  Horizontal brackets
are included by $c\in\mathcal I$, and smooth coefficients cause only
the displayed terms and horizontal multiples.  If all generators of
$\mathcal I$ vanished at $\theta_0$, $\mathcal N(\theta_0)$ would
have dimension at most two, contradicting Lie rank three.  Thus a
finite least order $m$ exists.

For $m\geq2$, interchange adjacent horizontal operators in an
$m$-word applied to $c$.  Its change is an outer horizontal word
applied to $\pm c\partial_\eta(H_{\mathrm{inner}}c)$.  The outer
word has length at most $m-2$.  Every Leibniz term contains a first
$c$ factor acted on by at most $m-2$ horizontal derivatives, which
vanishes at $\theta_0$ by minimality.  All adjacent interchanges
therefore preserve the evaluation.  Orders zero and one are
immediate.  The nonzero leading tensor is symmetric; polarization
then shows that its diagonal polynomial $P_m$ is nonzero.
\end{proof}

\subsection{Proof of Lemma~\ref{bridge:rectangles}}\label{release:proof23}
\begin{proof}
Choose independent unit vectors $e^{(1)},e^{(2)}$ avoiding the zeros
of $P_m$.  Their negatives also avoid these zeros by homogeneity.
Denote the four directions by $\mathcal D$ and let $e^\perp$ be the
counterclockwise unit rotation of $e$.  Since
$\det(e,e^\perp)=1$, $[H_e,H_{e^\perp}]=c\partial_\eta$.

Let $\Phi_X^t$ denote the local flow of $X$.  The lift of the base
rectangle
\[
 0\longrightarrow\varepsilon e
 \longrightarrow\varepsilon e+\delta e^\perp
 \longrightarrow\delta e^\perp\longrightarrow0
\]
has endpoint
\[
 h_e(\delta)=\Phi_{H_{e^\perp}}^{-\delta}
   \Phi_{H_e}^{-\varepsilon}
   \Phi_{H_{e^\perp}}^\delta
   \Phi_{H_e}^{\varepsilon}(\theta_0).
\]
It lies in the base-zero fiber, is smooth in signed $\delta$, and
$h_e(0)=\theta_0$.  Differentiating at zero gives the vertical vector
\[
 \Delta_e(\varepsilon)=
  (\Phi_{H_e}^{-\varepsilon})_*
       H_{e^\perp}(\Phi_{H_e}^{\varepsilon}(\theta_0))
       -H_{e^\perp}(\theta_0).
\]
For the iterated bracket $\operatorname{ad}_{H_e}^jH_{e^\perp}$,
write its vertical coefficient as $f_j$.  Then
$f_1=c$ and $f_{j+1}=H_e f_j-f_j\partial_\eta A_e$, where $A_e$
is the vertical coefficient of $H_e$.  Induction yields
\[
 f_j=H_e^{j-1}c+
       \sum_{\ell=0}^{j-2}a_{j\ell}H_e^\ell c
\]
for smooth coefficients $a_{j\ell}$.  Minimality of $m$ kills all
lower terms at $\theta_0$.  The finite Taylor formula for the smooth
flow pullback therefore gives
\[
 \Delta_e(\varepsilon)
  =\frac{\varepsilon^{m+1}}{(m+1)!}
       P_m(e)\partial_\eta+o(\varepsilon^{m+1}).
\]
This uses a finite Taylor remainder, not convergence of a smooth
Taylor series.  One sufficiently small positive $\varepsilon$ makes
all four derivatives nonzero.  The one-dimensional inverse function
theorem makes each signed $\delta$ family cover an open fiber
interval around zero.  Their finite intersection contains a common
open interval $I$.  Shrink the $\delta$ ranges, $I$, and all state
boxes so the finitely many loop families stay inside the ordinary and
$K_{\mathrm{loc}}$ domains.
\end{proof}

\subsection{Proof of Lemma~\ref{bridge:planar}}\label{release:proof24}
\begin{proof}
For each $b\ne0$, at least one $e\in\mathcal D$ satisfies $e\cdot b<0$,
because the two chosen axes are independent.
The rectangle lies in $e\cdot w\geq0$, for either sign of $\delta$;
the tail lies in $e\cdot w<0$ except at zero.  For $0<\rho<1$,
follow its first three edges and the returning fourth edge only to
$\rho\delta e^\perp$, join that point to $\rho b$ by a segment,
then follow the remaining tail to $b$.  The connector has negative
$e$-projection after its first point, so it cannot meet the rectangle.
It meets the tail only at $\rho b$, since $b$ is not parallel to
$e^\perp$.  The full polygon is consequently injective and still
starts exactly at zero.

For an explicit velocity estimate, assign each original rectangle
edge and the tail one unit of time.  Cut the return at time
$4-\rho$, traverse the connector until time $4+\rho$, and then
resume the original tail.  The connector velocity is
$(b-\delta e^\perp)/2$.  All changes occur on an interval of length
$2\rho$, giving $L^1$ error at most
$C\rho(\|b\|+|\delta|)$.  Round finitely many corners in mutually
disjoint shrinking disks.  Elementary smooth rounding of these
nondegenerate polygonal corners preserves injectivity, endpoints,
and regularity, and adds a vanishing $L^1$ velocity error.  The disks
may depend on the fixed $b,\delta$; no uniform radius over these
parameters is required.  The total variations are bounded.

If $b=0$, truncate the final return at $\rho\delta e^\perp$ and
round its corners.  On a fixed last time interval one can reduce the
return velocity by the factor $1-\rho$; the same $L^1$ conclusion
holds and the terminal base point tends to zero.  The case
$\delta=0$ is obtained later by continuity in $\delta$, without
approximating a degenerate out-and-back path by embedded arcs.

For the lift equation $\theta'=H_1(\theta)u'+H_2(\theta)v'$,
bounded state derivatives on a compact chart give
\[
 \sup_{t\leq T}\|\theta_j(t)-\theta(t)\|
 \leq C\|w_j'-w'\|_{L^1}
       \exp\!\left(C\int_0^T\|w_j'(t)\|\,dt\right).
\]
The bounded variations and vanishing $L^1$ error thus give uniform
convergence of lifts from the same $\theta_0$, and continuation keeps
the fine approximants inside the ordinary neighborhood.  Each
embedded regular represented lift is realized by one global smooth
loss using the full compatible jet construction in
Lemma~\ref{lem:jet}.  Ordinary closure holds at its
endpoint.  Passing to the endpoint limit through the already smooth
$K_{\mathrm{loc}}$ proves the required equality at every rectangle
and tail endpoint, including closed-loop endpoints.  Those limiting
closed loops are not asserted to be nonstationary autonomous loss
trajectories themselves.  Continuity of $h_e(\delta)$ and of its
straight-tail transport covers $\delta=0$.
\end{proof}

\subsection{\texorpdfstring{Proof of Theorem~\ref{bridge:main} for $q=2,p=3$}{Proof in the specified smooth regime}}\label{release:proof25}
\begin{proof}[Proof of Theorem~\ref{bridge:main} for $q=2,p=3$]
Let $P_b$ be horizontal transport along the straight base segment
from zero to $b$, starting at an arbitrary initial fiber coordinate
$\eta\in I$.  For every $b\ne0$ select an exterior direction $e$
and a signed $\delta$ such that $h_e(\delta)$ is precisely that
fiber point.  Lemma~\ref{bridge:planar} gives
\[
 M(P_b(\eta))=K_{\mathrm{loc}}(\phi(P_b(\eta))).
\]
For $b=0$ the same follows from the truncated-return limit.  The map
$\Psi(b,\eta)=P_b(\eta)$ is smooth, has base projection $b$, and
has an invertible fiber derivative at $(0,0)$.  The inverse function
theorem therefore makes it a local diffeomorphism from a smaller base
disk times fiber interval onto a plaque neighborhood of $\theta_0$.
The equality holds at every such point.  Discontinuous choice of a
witnessing direction $e$ is harmless: $\Psi$ itself is smooth, and
the equality is pointwise for every $(b,\eta)$.

Finally shrink the original ambient Frobenius product box so its
entire initial level lies in this neighborhood.  Extend
$K_{\mathrm{loc}}$ globally by a target cutoff equal to one near
$\phi(\theta_0)$, and shrink once more into that equality region.
This proves the metric property on the actual local invariant level
and completes Theorem~\ref{bridge:main}.
\end{proof}

\subsection{Proof of Proposition~\ref{bridge:flatness}}\label{release:proof26}
\begin{proof}
The stationary case is immediate.  Otherwise fix one open clock ball
$B$ and a state box inside the domain of $K_{\mathrm{loc}}$.  For
any fixed finite sequence $g_1,\ldots,g_k\in B$, let $E(t)$ be the
flow-word endpoint at durations $t_1,\ldots,t_k$.  All sufficiently
small positive duration vectors have a strict represented clock.
Smooth control ramps and the compatible autonomous jet construction
give genuine single-global-loss endpoints tending to $E(t)$.  The
smooth representative permits the limit, so
$R(E(t))=0$ throughout the positive open duration orthant near zero.

The finite endpoint map extends smoothly to a full neighborhood of
zero durations.  Every mixed derivative of $R\circ E$ on the
positive orthant is zero and remains zero at its corner by
continuity.  Differentiating once in each duration yields
\[
 Y_{g_1}\cdots Y_{g_k}R(\theta_0)=0,
 \qquad Y_g=-D\phi^{\mathsf T}g,
\]
with the naming order fixed by pullback convention; every sequence
order is available.  The zero-order identity follows at the initial
time.  Negative durations are used only to define derivatives of a
smooth map, not as assumed test trajectories.

For each fixed $k$ the displayed evaluation is multilinear in its
constant controls.  Vanishing on $B^k$ extends it to all controls,
and therefore to all ordered coordinate-gradient words.  Every
bracket field is a finite signed combination of such differential
operator words, and products of bracket fields retain this property.
Choose a finite local bracket frame $V_1,\ldots,V_p$ for
$\mathcal E$.  All its ordered products annihilate $R$ at
$\theta_0$.  Each plaque-coordinate vector field is a smooth linear
combination of this frame.  Repeated Leibniz expansion of any
coordinate derivative consequently consists of smooth coefficients
times ordered frame derivatives of $R$, all zero at $\theta_0$.
This proves flatness in every plaque chart.  The argument does not
identify a smooth flat function with zero on a neighborhood.
\end{proof}

\subsection{Proof of Proposition~\ref{bridge:hessian}}\label{release:proof27}
\begin{proof}
Since $q=2$, $X,Y$ smoothly frame $\mathcal H$.  Put
$\lambda=\langle X,Y\rangle=a^{\mathsf T}Mb$.  Euclidean
differentiation gives the exact identity
\[
 [X,Y]=\grad\lambda-2D^2u\,Y.
\]
The bracket lies in $\mathcal E$ and the last term is horizontal,
so the full ambient $\grad\lambda$ belongs to $\mathcal E$.
If ordinary closure holds, Lemma~\ref{lem:clock} gives a nonempty
plaque-open patch $O$ with the common full-matrix factorization.
Actual constant-rank fibers of $\phi|_L$ make $\lambda$ constant
along them.  Its gradient is therefore orthogonal to
$\mathcal E\cap\ker D\phi$.  Because it also lies in
$\mathcal E$, it belongs to $\mathcal H$, the orthogonal complement
of that vertical subspace inside $\mathcal E$.  The identity above
then makes $[X,Y]$ horizontal on $O$.

Smooth frame coefficients imply that the horizontal bundle is
involutive on this leaf-open patch.  Restriction of ambient tangent
fields to $L$ commutes with brackets.  All coordinate gradients are
smooth combinations of $X,Y$, so their generated distribution on
$O$ has rank two.  This contradicts constant $p>2$; an ambient-open
involutivity patch is not needed.  Conversely, when $p=q=2$, the
local recoverability argument already proves the metric property
and hence ordinary closure.  Finally $D^2u=0$ in the affine case and
$D^2u=\alpha\Id$ in the isotropic quadratic case, giving the stated
specializations without changing the Euclidean parameter metric.
\end{proof}

\section{Proofs of the autonomous-loss controls}

\subsection{Proof of Proposition~\ref{ctrl:delayed}}\label{release:proof28}
\begin{proof}
Write $F(x)=x^k$. The coordinate gradients are
\[
 X=\partial_x,\qquad Y=(F'(x)z,1,F(x)).
\]
They are independent everywhere and
$\operatorname{ad}_X^kY=k!\partial_z$, so they generate a rank-three
distribution everywhere. At the origin $Y-\partial_y$ has polynomial
coefficients homogeneous of degree $k$, independent of $y$. Every nonzero
bracket of length $\ell\ge2$ contains at least one such nonconstant factor
and at most $\ell-1$ coefficient differentiations, hence vanishes at the
origin when $\ell\le k$. Thus the stated delay is exact. The mobility is
\[
 M=\begin{bmatrix}
 1&F'z\\ F'z&1+F^2+(F'z)^2
 \end{bmatrix}.
\]
Its first parameter derivative is zero at the origin for $k\ge2$.
Consequently $\rank D(\phi_k,M)(0)=2$, so this necessary first-order test
alone does not detect the failure below.

Fix $a>0$ and $\epsilon>0$, and set, for all real $u$,
\begin{align*}
 h(u)&=\epsilon u(a-u),\\
 \zeta(u)&=\frac{1}{\sqrt{1+F(u)^2}}
       \int_0^u\frac{F(s)h'(s)}{\sqrt{1+F(s)^2}}\,ds,\\
 B(u)&=\frac{F'(u)\zeta(u)-h'(u)}{1+F(u)^2},\\
 A(u)&=\int_0^u\left[-1-
              \frac{(h'(s)-F'(s)\zeta(s))^2}{1+F(s)^2}\right]ds.
\end{align*}
All these functions are globally smooth, since the denominators are strictly
positive. Take
\[
 f_1(u,v)=-u,\qquad f_2(u,v)=A(u)+B(u)(v-h(u)).
\]
The first trajectory from the origin is $(x,y,z)=(t,0,0)$.
The second is
\[
 x=t,\qquad z=\zeta(t),\qquad y=h(t)-F(t)\zeta(t),
 \qquad \phi_k=(t,h(t)).
\]
Indeed, differentiating the defining integral gives
$\zeta'=F(h'-F'\zeta)/(1+F^2)=-FB$.
Along the displayed graph $\grad f_2=(A'-Bh',B)$, and the factor equations
are
\[
 \dot x=-A'+B(h'-F'\zeta)=1,\qquad
 \dot z=-FB=\zeta',\qquad \dot y=-B.
\]
The asserted derivative of $y$ equals
$(h'-F'\zeta)/(1+F^2)=-B$ as well. Thus uniqueness verifies the actual
autonomous trajectories, not merely a controlled proxy.

At time $a$ both outputs equal $(a,0)$. Integration by parts gives
\[
 \zeta(a)\sqrt{1+F(a)^2}
   =-\int_0^a\frac{h(s)F'(s)}{(1+F(s)^2)^{3/2}}\,ds<0.
\]
The first trajectory has $M_{12}=0$ and the second has
$M_{12}=F'(a)\zeta(a)<0$. They cannot share one entire mobility $K$ at the
common output. As $a\downarrow0$ with $\epsilon$ fixed, both parameter
paths on $[0,a]$ converge uniformly to the origin. Hence the witness lies
inside every proposed ordinary neighborhood for a small enough $a$.
This proves failure of \textup{(O)} with the exact required quantifiers.
\end{proof}

\subsection{Proof of Proposition~\ref{release:unlabelled1}}\label{release:proof29}
\begin{proof}
Here $X=(1,0,0,0)$ and $Y=(z,1,x+w,z)$ are independent;
$[X,Y]=\partial_z$ and $[\partial_z,Y]=\partial_x+\partial_w$.
These four vectors span $\R^4$ everywhere. The first output is affine,
so Proposition~\ref{bridge:hessian} excludes ordinary closure. Directly,
\[
 M=\begin{bmatrix}1&z\\z&1+(x+w)^2+2z^2\end{bmatrix},\qquad
 v=(0,-(x+w),1,0)\in\ker D\phi
\]
has $D M_{12}[v]=1$, violating the initial-plaque flatness consequence.
This example satisfies the rank hypotheses but not the universal ordinary
premise; it is not a counterexample to the equivalence in its proved regimes.
\end{proof}

\section{Proofs for restricted loss families}

\subsection{Proof of Proposition~\ref{loss:linear}}\label{release:proof30}
\begin{proof}
Let $E(v)$ be the time-one endpoint from $\theta_0$ of
$\dot\theta=-D\phi(\theta)^Tv$. For $v$ in a sufficiently small ball,
small-field local ODE theory gives existence on $[0,1]$ and smooth
dependence on $v$. At $v=0$,
\[
 E(0)=\theta_0,\qquad DE(0)=-D\phi(\theta_0)^T .
\]
Put $z_0=\phi(\theta_0)$ and $\Psi(v)=\phi(E(v))$. Then
$D\Psi(0)=-M_0$ is invertible. Shrinking to a ball $B_\delta(0)$,
the inverse function theorem makes $\Psi$ a diffeomorphism onto an
open target neighborhood $V$. Define
\[
 K_{\rm loc}(z)=M\bigl(E(\Psi^{-1}(z))\bigr),\qquad z\in V .
\]
Full row rank persists on a parameter neighborhood $U_0$, on which
$M\ge m\Id$ for some $m>0$. Choose a smaller $U\ni\theta_0$ with
closure in $U_0$, with $\phi(U)$ contained in a target neighborhood
$V_1$ whose closure is compact in $V$, and with
$|\phi(\theta)-z_0|<r$ on $U$, where $r/m<\delta$.

For $g\ne0$ let $n=g/|g|$. As long as its actual flow remains in $U$,
\[
 \frac{d}{dt}\bigl[n^T(\phi(\theta(t))-z_0)\bigr]
 =-|g|\,n^TM(\theta(t))n\le-m|g|.
\]
Integration and the target-radius bound imply $t|g|<r/m<\delta$.
Time rescaling of the constant-control vector field gives
$\theta(t)=E(tg)$. Thus $tg$ belongs to the inverse chart and
\[
 tg=\Psi^{-1}(\phi(\theta(t))),\qquad
 M(\theta(t))=K_{\rm loc}(\phi(\theta(t))).
\]
This argument covers the whole interval before exit, including arbitrarily
large $g$; it does not assume that $tg$ remains small without proving it.
For $g=0$ the solution is stationary.
A smooth target cutoff equal to one near $\overline V_1$ and supported
in $V$ extends $K_{\rm loc}$ by zero to a global smooth $K$.
No positivity outside the tested image is required.
\end{proof}

\subsection{Proof of Theorem~\ref{loss:quadratic}}\label{release:proof31}
\begin{proof}
Write $v=y+x^kz$. The exact parameter gradient equations are
\begin{align*}
 \dot x&=b_1-x-\gamma v
          +kx^{k-1}z(b_2-\gamma x-v),\\
 \dot y&=b_2-\gamma x-v,\qquad
 \dot z=x^k(b_2-\gamma x-v).
\end{align*}
For one fixed $c>0$, introduce
\[
 s=\frac{t}{ca},\quad x=aX,\quad y=a^2Y,\quad z=a^{k+2}Z,
 \qquad b_1=B/c,\quad b_2=aC .
\]
Put $V=Y+a^{2k}X^kZ$, so that $v=a^2V$. On $0\le s\le1$
the exact scaled initial-value problem, starting at zero, is
\begin{align}
 X'&=B-caX-c\gamma a^2V
       +ck a^{2k+2}X^{k-1}Z(C-\gamma X-aV),\label{loss:scaled-X}\\
 Y'&=c(C-\gamma X-aV),\label{loss:scaled-Y}\\
 Z'&=cX^k(C-\gamma X-aV).\label{loss:scaled-Z}
\end{align}
These equations are polynomial in states and parameters $a,B,C$.
At $a=0$ the solutions on $[0,1]$ are bounded polynomials.
Local ODE dependence therefore gives existence and analytic dependence
on the whole interval for parameters in a small neighborhood of the
reference values. Concretely, choose a compact tube around the limiting
trajectory; the integral equation and a Gronwall estimate prevent exit
for nearby parameters. No global completeness of polynomial vector
fields is asserted.

Let $E_c(a,B,C)=(X(1),V(1))$. At $a=0$,
\[
 E_c(0,B,C)=\bigl(B,c(C-\gamma B/2)\bigr).
\]
At $(B,C)=(1,\gamma/2)$ the endpoint is $(1,0)$, and the shooting
Jacobian is
\[
 D_{(B,C)}E_c(0,1,\gamma/2)
 =\begin{pmatrix}1&0\\-c\gamma/2&c\end{pmatrix},
 \qquad \det D_{(B,C)}E_c=c>0 .
\]
The analytic implicit function theorem supplies $B_c(a),C_c(a)$ with
$B_c(0)=1$, $C_c(0)=\gamma/2$ and
$E_c(a,B_c(a),C_c(a))=(1,0)$ exactly.
Thus $b_c(a)=(B_c(a)/c,aC_c(a))$ gives the exact represented endpoint
$(a,0)$. Use a common sufficiently small $a_*$ for $c_1,c_2$.
The equality is an actual endpoint identity, not a formal-series
substitute for a shooting solution.

At $a=0$ and these limiting shooting parameters, $X(s)=s$ and
\[
 Z(1)=c\gamma\int_0^1s^k(1/2-s)\,ds
      =-\frac{c\gamma k}{2(k+1)(k+2)} .
\]
Smooth dependence in~\eqref{loss:scaled-Z} yields
\[
 z(ca)=-\frac{c\gamma k}{2(k+1)(k+2)}a^{k+2}
          +O(a^{k+3}).
\]
The full mobility is
\[
 M(x,y,z)=
 \begin{pmatrix}
 1&kx^{k-1}z\\
 kx^{k-1}z&1+x^{2k}+(kx^{k-1}z)^2
 \end{pmatrix}.
\]
At each exact endpoint $x=a$, multiplication by $ka^{k-1}$
and subtraction give~\eqref{loss:gap}. The constants implicit in the
remainders depend on the fixed $k,\gamma,c_1,c_2$.
A nonzero leading coefficient implies inequality for all sufficiently
small positive $a$.

The scaled states stay uniformly bounded on $[0,1]$ as $a\to0$.
Consequently $x=O(a)$, $y=O(a^2)$ and $z=O(a^{k+2})$ uniformly on
the whole actual trajectories. They therefore fit inside every proposed
common neighborhood for small enough $a$.
A single $K$ cannot take both unequal mobility values at the same
output $(a,0)$. This is precisely the required failure of the restricted
ordinary-closure property. Its quantifier permits different horizons.
\end{proof}

\subsection{Proof of Corollary~\ref{loss:squared}}\label{release:proof32}
\begin{proof}
For $\tau=W^{-1}b$,
\[
 \ell_\tau(\psi_k(\theta))
   =f_b(\phi_k(\theta))+\tfrac12 b^TH^{-1}b .
\]
The additive constant changes no Euclidean parameter gradient field.
Thus the two actual paths in Theorem~\ref{loss:quadratic} are also
paths for ordinary squared error on this one fixed representation,
and their represented endpoints both equal $W(a,0)^T$.
Moreover
\[
 D\psi_k=W D\phi_k,\qquad
 M_{\psi_k}=W M_{\phi_k}W^T.
\]
Invertibility of $W$ preserves the nonzero full mobility difference;
a particular matrix entry need not be preserved.
The coordinate gradients are invertible constant linear combinations
of those of $\phi_k$, so their generated Lie distributions are the same.
The ranks follow from Remark~\ref{loss:linear-separation}.
The formula for $M_{\phi_k}$ gives $DM_{\phi_k}(0)=0$ for $k\ge2$;
constant congruence gives the same property for $M_{\psi_k}$.
\end{proof}

\subsection{Proof of Proposition~\ref{loss:regularity}}\label{release:proof33}
\begin{proof}
At $b=0$ the parameter flow from $\theta_0$ is stationary.
Let $J_0=D\phi(\theta_0)$ and $M_0=J_0J_0^T$.
At one fixed sufficiently short time $\tau>0$, smooth flow dependence
defines the endpoint for all sufficiently small $b$, with the entire
path in the ordinary neighborhood.
If $P(t)$ is the derivative of the parameter state with respect to $b$
at zero, variation of the vector field gives
\[
 \dot P=-J_0^TJ_0P-J_0^T,\qquad P(0)=0 .
\]
Consequently $Q(t)=J_0P(t)$ satisfies
$\dot Q=-M_0Q-M_0$, $Q(0)=0$, so
\begin{equation}\label{loss:section-derivative}
 D_b\phi(\theta_b(\tau))\big|_{b=0}
       =-(\Id-e^{-\tau M_0}).
\end{equation}
This identity does not require $M_0$ to be invertible.
On $S=\operatorname{im}M_0$, its eigenvalues are
$-(1-e^{-\tau\lambda})$ for $\lambda>0$, hence nonzero.
Restrict $b$ to the $q$-dimensional space $S$.
The constant-rank theorem supplies a local embedded image of $\phi$
with tangent $S$ at $z_0$, after shrinking the parameter neighborhood.
In a chart of that image,~\eqref{loss:section-derivative} is an invertible
$q$-dimensional endpoint derivative.
The inverse function theorem produces a section $\sigma$ satisfying
$\phi(\sigma(z))=z$ on a represented neighborhood of $z_0$,
where each $\sigma(z)$ is an actual endpoint of one of the tested flows.
Thus
\[
 K(z)=M(\sigma(z))
\]
there, by the common-closure hypothesis.
The right side is smooth. A local target chart flattening the embedded
image extends it normally; a target cutoff then gives a global smooth
extension agreeing on a smaller image.
In the analytic case the local ODE, image charts and inverse theorem
give an analytic section and a local analytic extension; no global
analytic cutoff is asserted.
For $q<d$ the inverse theorem has only been used on the image, not on
all of $\R^d$.
\end{proof}

\section{Finite checks and reproducibility}
The universal statements above rest on the included proofs. Existing exact
algebraic checks are finite implementation evidence and were not rerun to
produce this manuscript. The latest boundary validation used rational
arithmetic on 39 fixtures: 24 effective scalar-imbalance and 15 nonscalar
cases, including four zero-scalar cases. It checked 259 hidden-rotation and
tangent identities and reported zero mismatches. These were bounded algebraic checks, not model training. These finite
checks neither prove the universal classification nor establish task accuracy,
statistical uncertainty, approximate-closure error or a practical speedup.

The local provenance audit preserves the available inputs, proof versions,
automated review records, primary source texts, and their hashes. Complete local source custody is distinguished from selected
theorem-and-proof reading. The two closest source versions are
arXiv:2508.07370v2 (12 March 2026, ICLR 2026 rendition) and
arXiv:2307.00144v2 (10 July 2024 revision of the NeurIPS 2023 work).
The following numerical record is copied from a frozen, actually executed
experiment. No new scientific run was used to assemble this manuscript.
The algebraic and autonomous-loss records answer different finite questions.

\section{A finite autonomous-loss implementation check}
\label{num:sentinel}
We executed one deterministic CPU sentinel for the engineered smooth-loss
construction in Proposition~\ref{ctrl:delayed}. This tests an implementation
of a proved mechanism, rather than a learning algorithm. The fixed-Hessian
quadratic and ordinary squared-loss extension in this manuscript has
\emph{not} been executed: its numerical $a_*$ and any executed quadratic-loss
grid remain unknown. The measurements below must not be transferred to that
new loss family.

\paragraph{Actual objective and parameter flow.}
Set $F(u)=u^k$, $h(u)=u(a-u)$ and $D(u)=1+F(u)^2$. The implemented global
objective was
\[
 f_2(u,v)=A(u)+B(u)(v-h(u)),\qquad
 B=\frac{F'\zeta-h'}{D},\qquad
 A'=-1-\frac{(h'-F'\zeta)^2}{D},\quad A(0)=0,
\]
where
\[
 \zeta(u)=D(u)^{-1/2}\int_0^u F(s)h'(s)D(s)^{-1/2}\,ds.
\]
Writing $v=y+F(x)z$, its full output gradient and actual parameter ODE were
\begin{align*}
 g_u&=A'+B'(v-h)-Bh',&g_v&=B,\\
 (\dot x,\dot y,\dot z)&=-(g_u+F'(x)zB,\ B,\ F(x)B).
\end{align*}
In particular the off-graph term $B'(v-h)$ was included; substituting the
analytic graph for this ODE would not qualify the implementation.
The analytic comparison path is
$(t,h(t)-F(t)\zeta(t),\zeta(t))$. The comparison loss $f_1(u,v)=-u$
was also integrated, with analytic path $(t,0,0)$.

\paragraph{Frozen protocol and discriminating controls.}
The twelve witness cases used $k\in\{2,3,4\}$ and
$a\in\{0.5,0.35,0.25,0.175\}$, with $\epsilon=1$. Four matched controls
used $F\equiv0$ at the same $a$ values; these have $p=q=2$, mobility $I_2$
and zero gap. We used DOP853 with relative tolerance $2\times10^{-12}$,
absolute tolerance $10^{-14}$ and 21 stored times per trajectory.
Working scalar quadratures used absolute and relative tolerances $10^{-13}$
and at most 100 subdivisions. Independently, a 50-decimal-digit quadrature
evaluated the cancellation-avoiding reference
\begin{equation}\label{num:reference}
 \Delta_{\rm ref}=
 -\frac{k^2a^{2k+1}}{\sqrt{1+a^{2k}}}
 \int_0^1\frac{t^k(1-t)}{(1+a^{2k}t^{2k})^{3/2}}\,dt.
\end{equation}
This is a high-precision estimate, not a certified quadrature interval.

Five-point central differences, step $2\times10^{-4}$, differentiated the
globally defined scalar loss composed with $\phi_k$ in all three parameter
coordinates. Both probes were off graph: $(x,y,z)=(0,0.1,0.03)$, and
$x=0.3a$, $z=0.02a$, $v=h(x)+0.05a$, with $y=v-F(x)z$.
Each coordinate used tolerance $5\times10^{-8}+5\times10^{-6}|g_{\rm FD}|$.
The deliberately incorrect gradient omitting $B'(v-h)$ had to fail the same
tolerance at both probes. Its first-probe discrepancy had to exceed $10^{-3}$.
This matters because the omitted term vanishes on the analytic graph.

Before execution we fixed state error $\le3\times10^{-10}$, represented
endpoint error $\le10^{-10}$, nonzero-gap relative error $\le5\times10^{-6}$,
and a zero-control gap threshold $10^{-12}$. For
$L=k^2a^{2k+1}/((k+1)(k+2))$, the measured $-\Delta/L$ was compared with
$[(1+a^{2k})^{-2},(1+a^{2k})^{-1/2}]$ using slack $5\times10^{-6}$.
Nonfinite outputs could not pass. The first completed failed case or exception
would stop the sole attempt; no grid or tolerance was selected after observing
results. There are no random seeds or statistical confidence intervals.

\begin{table}[htbp]
\centering
\caption{Measured endpoint mobility gap
$\Delta=M_{12}(f_2;a)-M_{12}(f_1;a)$ and relative discrepancy
$|\Delta-\Delta_{\rm ref}|/|\Delta_{\rm ref}|$.
All twelve frozen engineered-loss witnesses are shown.}
\label{num:gaps}
\begin{tabular}{rrrr}
\toprule
$k$ & $a$ & Measured $\Delta$ & Relative discrepancy\\
\midrule
2 & 0.500 & $-9.9091061\times10^{-3}$ & $2.96\times10^{-13}$\\
2 & 0.350 & $-1.7294234\times10^{-3}$ & $8.12\times10^{-14}$\\
2 & 0.250 & $-3.2447983\times10^{-4}$ & $1.49\times10^{-12}$\\
2 & 0.175 & $-5.4668171\times10^{-5}$ & $2.82\times10^{-12}$\\
3 & 0.500 & $-3.4737275\times10^{-3}$ & $6.98\times10^{-12}$\\
3 & 0.350 & $-2.8911620\times10^{-4}$ & $2.17\times10^{-11}$\\
3 & 0.250 & $-2.7460640\times10^{-5}$ & $4.21\times10^{-11}$\\
3 & 0.175 & $-2.2618782\times10^{-6}$ & $5.40\times10^{-11}$\\
4 & 0.500 & $-1.0386359\times10^{-3}$ & $2.89\times10^{-12}$\\
4 & 0.350 & $-4.2027935\times10^{-5}$ & $1.99\times10^{-12}$\\
4 & 0.250 & $-2.0344820\times10^{-6}$ & $4.54\times10^{-13}$\\
4 & 0.175 & $-8.2099570\times10^{-8}$ & $2.26\times10^{-13}$\\
\bottomrule
\end{tabular}
\end{table}

\paragraph{Observed results and limits.}
All twelve witnesses and four matched controls passed the frozen criteria.
The maximum state discrepancy over the 21 stored times was $2.015\times10^{-12}$ and represented
endpoint discrepancy $1.859\times10^{-13}$; all four $F=0$ gaps were stored
as zero and their identity-mobility checks passed. All 32 off-graph probes
passed, with maximum coordinate FD discrepancy $1.043\times10^{-13}$.
Every omitted-term mutant was rejected; the minimum across probes of its
largest coordinate error was $0.0175$.
The $k=4,a=0.175$ ratio band has width $1.3195\times10^{-6}$, smaller than
the allowed slack, and is explicitly tolerance dominated. Passing that band
does not demonstrate sharp-band or next-order asymptotic resolution.
The integrator tolerances are not global error certificates, and working
quadrature error estimates are heuristic.

The historical implementation used 3122 $f_2$ right-hand-side evaluations
and 9057 quadrature calls. No GPU computation or model training was used.
Raw trajectories, gradient probes, numerical references, code/input hashes are preserved in the local historical record.
The supplied numerical payload can be run independently with the listed
dependencies; two original external resource-wrapper files are unavailable,
so the historical execution environment and resource measurements are not
fully reproducible from this package. The public historical data consist of
the exported summary and trajectory CSVs; the original raw probe/reference
JSON files and source-custody receipts are not distributed in this package.
These finite observations support implementation and mechanism agreement;
the exact endpoint equality and universal quantifiers rest on the proofs.
They establish no learning improvement, algorithmic speedup, numerical
validation of the new quadratic theorem, or general asymptotic law.

\paragraph{Separate pre-submission replay.}
A separate pre-submission replay under Python 3.12.14, NumPy 2.3.5,
SciPy 1.17.0, and mpmath 1.3.0 passed all 16 cases and 32 off-graph probes,
rejecting every omitted-term mutant. It used 3116 $f_2$ right-hand-side
evaluations and 7035 quadrature calls. Source copies, outputs, environment,
and comparisons are in \texttt{anc/audit\_replay/}. These new results are
separate from the historical measurements; they do not reproduce the
historical environment or resource wrappers, and do not test the
fixed-Hessian quadratic theorem.

\end{document}